\documentclass[letterpaper]{article} 
\usepackage{aaai25}  
\usepackage{times}  
\usepackage{helvet}  
\usepackage{courier}  
\usepackage[hyphens]{url}  
\usepackage{graphicx} 
\usepackage{natbib}  
\usepackage{caption} 
\usepackage{algorithm}
\usepackage[noend]{algpseudocode}
\usepackage{amsmath,amsfonts,amsthm}
\usepackage{adjustbox}
\usepackage{textcomp}
\usepackage{subcaption}
\usepackage{tabularx}
\usepackage{booktabs}
\usepackage{multirow}
\usepackage{makecell}
\usepackage{amssymb}
\usepackage{array}

\newtheorem{manualthminner}{Theorem}
\newenvironment{manualthm}[2][]{
  \renewcommand\themanualthminner{#2}
  \manualthminner[#1]
}{%
  \endmanualthminner
}

\newtheorem{manualleminner}{Lemma}
\newenvironment{manuallem}[2][]{
  \renewcommand\themanualleminner{#2}
  \manualleminner[#1]
}{%
  \endmanualleminner
}

\newtheorem{manualtheorem}{Theorem}

\newtheorem{manuallemma}{Lemma}

\newtheorem{manualcorollary}{Corollary}

\newtheorem{manualdefinition}{Definition}

\usepackage{newfloat}
\usepackage{listings}
\DeclareCaptionStyle{ruled}{labelfont=normalfont,labelsep=colon,strut=off} 
\floatstyle{ruled}
\newfloat{listing}{tb}{lst}{}
\floatname{listing}{Listing}
\title{Adaptive Branch Specialization in Spectral–Spatial Graph Neural Networks for Certified Robustness}

\author{
    Yoonhyuk Choi$^1$, Jiho Choi$^2$, Chong-Kwon Kim$^3$
}
\affiliations{
    \textsuperscript{1}Sookmyung Women's University, Seoul, Republic of Korea\\

    \textsuperscript{2}Korea Advanced Institute of Science and Technology (KAIST), Seoul, Republic of Korea\\
    
    \textsuperscript{3}Korea Institute of Energy Technology (KENTECH), Naju, Republic of Korea\\
    chldbsgur123@sookmyung.ac.kr, jihochoi1993@gmail.com, 
    ckim@kentech.ac.kr
}

\begin{document}

\maketitle

\begin{abstract}
Recent Graph Neural Networks (GNNs) combine spectral–spatial architectures for enhanced representation learning. However, limited attention has been paid to certified robustness, particularly regarding training strategies and underlying rationale. In this paper, we explicitly specialize each branch: the spectral network is trained to withstand $\ell_{0}$ edge flips and capture homophilic structures, while the spatial part is designed to resist $\ell_{\infty}$ feature perturbations and heterophilic patterns. A context-aware gating network adaptively fuses the two representations, dynamically routing each node’s prediction to the more reliable branch. This specialized adversarial training scheme uses branch‑specific inner maximization (structure vs feature attacks) and a unified alignment objective. We provide theoretical guarantees: (i) expressivity of the gating mechanism beyond 1-WL, (ii) spectral-spatial frequency bias, and (iii) certified robustness with trade-off. Empirically, SpecSphere attains state‑of‑the‑art node classification accuracy and offers tighter certified robustness on real‑world benchmarks. Our code is available at this \textit{link}\footnote{https://anonymous.4open.science/r/SpecSphere-684F}.
\end{abstract}

%

\section{Introduction} \label{sec:intro}
Graph neural networks (GNNs) have become the modern systems for many tasks such as node classification \cite{kipf2016semi}, link prediction \cite{zhang2018link}, and graph reasoning \cite{lee2019self}. Most popular GNN architectures implement Laplacian‑based filtering or message passing, implicitly enforcing a low‑pass prior that assumes homophily, where adjacent nodes share labels and features \cite{nt2019revisiting}. However, many real-world graphs exhibit heterophily in which connected nodes belong to different classes or display contrasting attributes \cite{pei2020geom, wang2024understanding}. Under these conditions, the inherent low-pass bias of standard graph convolutions smooths away critical high-frequency signals \cite{yan2021two}. Without explicit priors, even advanced spectral filters struggle to recover the lost discriminative details \cite{duan2024unifying}. Although recent studies combine spectral and spatial networks to boost representational power \cite{chen2023bridging,geisler2024spatio}, they do not exploit branch‑specialized training, which can improve robustness against adversarial attacks.

Minor structure edits $\ell_0$ (e.g., a handful of edge flips) can invert predictions \cite{ma2020towards}, while small $\ell_\infty$ perturbations on node features suffice to fool classifiers \cite{jin2021adversarial}. Although adversarial training enhances empirical robustness, it often overfits a single threat model and remains vulnerable to adaptive attacks \cite{gosch2023adversarial}. Certified defenses aim to provide formal worst-case guarantees, yet existing certification methods are predominantly limited to single-branch architectures and lack generalization capabilities \cite{zugner2020adversarial}. In addition, expressivity limits standard message‑passing GNNs to the power of the one‑dimensional Weisfeiler–Lehman (1‑WL) test. Extensions inject random features \cite{dwivedi2023benchmarking}, aggregate higher‑order substructures \cite{abu2019mixhop}, or adapt spectral responses \cite{bo2021beyond,duan2024unifying}. Nonetheless, no prior work unifies high expressivity with node‑wise homophily adaptation and a provable, dual‑norm robustness guarantee. 

Unlike prior spectral-spatial models that fuse branches via a static or purely data-driven gate, SpecSphere explicitly specializes its two branches and fuses them through a node-wise, context-aware gate. The spectral branch is regularized toward low-frequency and is adversarially trained against $\ell_{0}$ edge flips, while the spatial branch emphasizes high-frequency, heterophilic signals that are robust against $\ell_{\infty}$ feature perturbations. For robustness certification, we separately bound the Lipschitz constants of both branches, composed them through the fusion module, and coupled these bounds with classification margins. In this paper, we propose a spectral–spatial GNN that simultaneously achieves adaptive specialization and provable robustness as follows:
\begin{itemize}
    \item \textbf{Adaptive branch specialization and context-aware fusion.} A learnable gate, conditioned on local homophily and robustness signals, dynamically selects between a homophily/structure-robust spectral branch and a heterophily/feature-robust spatial branch.
    \item \textbf{Expressiveness beyond 1-WL.} A single spectral layer distinguishes the Cai–Fürer–Immerman (CFI) graphs, and the full network is universal for continuous permutation-equivariant functions.
    \item \textbf{Robustness certificates.} By composing branch-wise Lipschitz bounds with node-level margins through the fusion module, we certify each prediction against both $\ell_{0}$ edge flips and $\ell_{\infty}$ feature perturbations under real-world homophilic/heterophilic datasets.
\end{itemize}

\begin{table*}[!t]
\centering
\small
\begin{tabular}{|l|>{\centering\arraybackslash}p{1.8cm}|>{\centering\arraybackslash}p{2.4cm}|>{\centering\arraybackslash}p{1.8cm}|>{\centering\arraybackslash}p{1.8cm}|>{\centering\arraybackslash}p{2.5cm}|}
\hline
\textbf{Model} & \textbf{Architecture} & \textbf{Certification} & \textbf{Adaptivity} & \textbf{Expressivity} & \textbf{Branch Specialize} \\
\hline
Certified GNN \cite{wang2021certified} & Single & Certified $\ell_0$ & No & No & No \\
\hline
ACM‑GNN \cite{luan2022revisiting} & Hybrid & No & Adaptive & Standard & No \\
\hline
EvenNet \cite{lei2022evennet} & Spectral & Certified $\ell_\infty$ & Static & Standard & No \\
\hline
GARNET \cite{deng2022garnet} & Spectral & Empirical & Adaptive & Standard & No \\
\hline
RES-GRACE \cite{lin2023certifiably} & Single & Certified ($\ell_0$, $\ell_\infty$) & No & Standard & No \\
\hline
S$^2$GNN \cite{geisler2024spatio} & Hybrid & No & No & $>1$‑WL & No \\
\hline
TFE‑GNN \cite{duan2024unifying} & Spectral & No & Adaptive & Standard & No \\
\hline
PCNet \cite{li2024pc} & Dual & No & Adaptive & Standard & No \\
\hline
SPCNet \cite{li2025simplified} & Dual & Partial & Adaptive & Standard & No \\
\hline
\textbf{SpecSphere (ours)} & Dual & Certified ($\ell_0$, $\ell_\infty$) & Adaptive & $>1$‑WL & Yes \\
\hline
\end{tabular}
\caption{Comprehensive comparison of spectral–spatial and robustness‑aware GNNs. “Architecture” distinguishes single‑pass (spectral or spatial), hybrid (one filter type), and dual‑pass (both). “Adaptivity” indicates dynamic weighting/gating. “Branch Specialize” denotes explicit branch‑specific objectives/training. More baselines are introduced in \textbf{Appendix A}.}
\label{recent_methods}
\end{table*}

\section{Related Work} \label{sec:related}
\textbf{Homophily and Heterophily.} Laplacian message passing excels when nearby nodes share labels, as demonstrated by the seminal models \cite{kipf2016semi,velickovic2017graph}. However, in heterophilic graphs, this same low-pass bias blurs class boundaries. Both empirical and theoretical studies show that high heterophily shrinks prediction margins and exacerbates over-smoothing \cite{zhu2022does}. Consequently, recent methods inject high-frequency information through higher-order mixing \cite{abu2019mixhop}, separate aggregation schemes \cite{zhu2020beyond,lim2021large,choi2023signed}, or frequency-adaptive filters \cite{bo2021beyond,ko2023signed,duan2024unifying}. While these techniques boost performance on heterophilic graphs, they typically operate solely in either the spatial or spectral domain and do not address adversarial robustness.

\textbf{Spectral, Spatial, Hybrid, Dual‑Pass.} \textit{Spectral} GNNs design filters in the graph Fourier basis \cite{defferrard2016convolutional,bruna2013spectral}, whereas \textit{spatial} methods propagate information along edges \cite{hamilton2017inductive}. \textit{Hybrid} layers integrate both perspectives within a single operation, which includes heterophily-aware spectral–spatial kernels \cite{chien2020adaptive,he2024exploitation}. A complementary line of work employs adaptive mixing pipelines that split computation into two successive branches \cite{zhou2023dpgnn,chen2023bridging,li2024pc,shen2024dual,geisler2024spatio}. Although these architectures enhance expressivity and stabilize training, they still lack formal guarantees against adversarial graph attacks.

\textbf{Robust Graph Learning.} Empirical evidence indicates that higher heterophily degrades GNN robustness by reducing the number of adversarial edge flips required to manipulate predictions \cite{zhu2022does}. Although adversarial training enhances average-case defenses, it often overfits to specific perturbation budgets and remains vulnerable to adaptive attacks \cite{zugner2019adversarial}. Extensions based on randomized smoothing offer certified guarantees for single-branch models \cite{scholten2023hierarchical}, but they neither address heterophily nor extend to multi-branch architectures. Recently, \cite{lin2023certifiably} introduced the framework that certifies unsupervised graph contrastive learning representations via randomized edge‑drop smoothing. However, no spectral–spatial GNNs provide formal protection against both $\ell_0$ edge flips and $\ell_\infty$ feature perturbations \cite{gosch2023adversarial,hou2024robust}, underscoring a critical open challenge in robust graph learning.

\textbf{SpecSphere vs Existing Methods.} Table \ref{recent_methods} shows that SpecSphere is the first spectral–spatial GNN to offer both certified ($\ell_0$,$\ell_\infty$) robustness and a branch specialization. Unlike prior dual‑pass designs with only partial or no theoretical robustness guarantees, our method computes a closed‑form certificate directly from its learned margins and filter norms. In addition, the dual‑pass architecture dynamically interpolates between low‑ and high‑frequency signals across the entire homophily–heterophily spectrum. By combining this adaptive filtering with branch specialization and expressivity beyond 1‑WL, SpecSphere overcomes both the adversarial and expressive limitations of existing GNNs.

\section{Preliminaries}\label{sec:prelim}
Let $G=(V,E)$ be an undirected graph with $|V|=n$ nodes and $|E|=m$ edges. Let $A\in\{0,1\}^{n\times n}$ be the adjacency and $D=\operatorname{diag}(d_1,\dots,d_n)$ degree matrix, and
\begin{equation}
\label{eq_norm_lap}
\mathrm{
\mathcal{L}=I-D^{-1/2}AD^{-1/2}
}
\end{equation}
be the normalized Laplacian. $\mathcal{L}$ is symmetric positive‐semi‑definite with spectrum $0=\lambda_1\le\dots\le\lambda_n\le2$. Let $U=[u_1|\dots|u_n]$ collect its orthonormal eigenvectors; then $\mathcal{L}=U\Lambda U^{\top}$ with $\Lambda=\operatorname{diag}(\lambda_1,\dots,\lambda_n)$. The columns of $U$ form the graph Fourier basis, and the $\lambda_i$ are the corresponding frequencies \cite{grover2025spectro}. Node features are aggregated in $X\in\mathbb{R}^{n\times d_0}$, and class labels $Y\in\{1,\dots,C\}^n$ are given for labeled nodes (training subset) $V_L \subseteq V$.

\paragraph{Spectral filters.}
A $K$‑th order polynomial filter acts on $X$ is given by:
\begin{equation}
\label{eq:cheb_base}
\mathrm{
h_\alpha(\mathcal{L})X=\sum_{k=0}^{K}\alpha_k\mathcal{L}^kX,
}
\end{equation}
where $\alpha_0,\dots,\alpha_K\in\mathbb{R}$ are learnable coefficients. Equivalently, one may use a Chebyshev expansion $h_\alpha(\mathcal{L})=\sum_{k=0}^{K}\alpha_k T_k(\tilde{\mathcal{L}})$ with $\tilde{\mathcal{L}}=2\mathcal{L}/\lambda_{\max}-I$ under the usual normalization ($\lambda_{\max}=2$).

\paragraph{Spatial message passing.}
A generic message‑passing layer updates hidden states $H^{(\ell)}$ via
\begin{equation}
\mathrm{
H^{(\ell+1)}
  =\sigma \bigl(A^{(\ell)}H^{(\ell)}W^{(\ell)}\bigr),
  \qquad H^{(0)}=X,
}
\end{equation}
where $A^{(\ell)}$ stands for the learnable edge weights, $W^{(\ell)}$ is a node‑wise linear map, and $\sigma$ is a point‑wise non‑linearity.

\begin{manualdefinition}[1‑WL Equivalence]\label{def:1wl}
Two graphs are 1‑WL indistinguishable if the Weisfeiler–Lehman vertex refinement assigns the same colored multiset at every iteration. Standard GNNs are at most 1‑WL \cite{wijesinghe2022new}.
\end{manualdefinition}

\begin{manualdefinition}[Local Homophily]
We estimate a continuous homophily score $\mathcal{H}_i \in [0,1]$ for node $i$, where $\mathcal{H}_i \approx 1$ means a highly homophilic neighborhood and $\mathcal{H}_i \approx 0$ indicates heterophilic surroundings. One simple estimator is:
\begin{equation}
\label{def:homo}
\mathcal{H}_i = \frac{1}{|\mathcal{N}(i)|} \sum_{v\in\mathcal{N}(i)} \mathrm{1} \bigl[\hat{y}_i = \hat{y}_v\bigr],
\end{equation}
where $\hat{y}_i$ is a pseudo-label or the current prediction.
\end{manualdefinition}

\begin{manualdefinition}[Permutation Equivariance]\label{def:eq}
Let $\Pi$ be an $n\times n$ permutation matrix. A node‑level operator $F$ is equivariant if $F(\Pi X,\Pi A\Pi^{\top})=\Pi F(X,A)$ for all $\Pi$. Graph‑level operators are invariant when the right‑hand side drops the leading $\Pi$. Both spectral and spatial operations used in SpecSphere satisfy this property.
\end{manualdefinition}


\section{Methodology}\label{sec:model}
SpecSphere consists of two complementary branches. Each branch is further trained to defend against a different type of perturbation (structure vs feature attacks). Then, a node-wise adaptive fusion gate decides which branch to trust more based on local homophily and robustness indicators. We theorize how this specialization and fusion pipeline improves expressivity and robustness. Note that the $\boxed{\text{boxed equations}}$ result in a total loss in Eq. \ref{full_obj}.

\subsection{Spectral Branch (Structural Robustness)}
\label{sec:spectral}
Let $\mathcal{L}$ be the normalized Laplacian (Eq. \ref{eq_norm_lap}) and define $\tilde{\mathcal L} = \mathcal L - I$. Expanding a polynomial filter in the Chebyshev basis (Eq. \ref{eq:cheb_base}), we implement the spectral branch as
\begin{equation}
  f_{\mathrm{spec}}(X;\Theta_{\mathrm{spec}})
   = 
  \sum_{k=0}^{K}
    T_k \bigl(\tilde{\mathcal L}\bigr) X \Theta_k,
\end{equation}
where $\{T_k\}$ are Chebyshev polynomials and $\Theta_{\mathrm{spec}}$ are trainable parameters. Layer stacking and low-pass regularization can be achieved by setting $H_{\text{spec}}^{(0)}=X$ as follows:
\begin{equation}
  H_{\text{spec}}^{(\ell)} = \sigma \left(
      \sum_{k=0}^{K}
        T_k(\tilde{\mathcal L}) 
        H_{\text{spec}}^{(\ell-1)} 
        \Theta_k^{(\ell)}
    \right),
    \quad \ell=1,\dots,L_s.
  \label{eq:spec-layer-cheb}
\end{equation}
To explicitly bias the spectral branch toward smooth (homophilic) signals and structure robustness, we add a low-pass regularizer:
\begin{equation}
\boxed{\mathcal{R}_{\text{LP}} =
\sum_{\ell=1}^{L_s}
\left\|
  \mathcal{L}^{1/2} H_{\text{spec}}^{(\ell)}
\right\|_F^2}
\end{equation}
which penalizes high-frequency energy in $H_{\text{spec}}^{(\ell)}$. The final output is $Z_{\text{spec}} := H_{\text{spec}}^{(L_s)} \in \mathbb{R}^{n\times d_\ell}$. Any off-the-shelf spectral GNN can be substituted here to improve expressivity.

\subsection{Spatial Branch (Feature Robustness)}
\label{sec:spatial}
With $H_{\text{spat}}^{(0)} = X$, the generic GAT-style spatial layer is:
\begin{equation}
  H_{\text{spat}}^{(\ell)} 
  =
  \sigma  \bigl(
    \hat{A} H_{\text{spat}}^{(\ell-1)}W_{\text{spat}}^{(\ell)}
  \bigr),
  \quad \ell=1,\dots,L_p,
  \label{eq:spatial-layer}
\end{equation}
where $\hat{A}$ is obtained by a feature-based attention mechanism. For strongly heterophilic graphs, we optionally use specialized spatial modules such as FAGCN \cite{bo2021beyond}:
\begin{align}
\begin{aligned}
F_{\mathrm{LP}}^{(\ell)} &= \hat{A} H_{\text{spat}}^{(\ell-1)} W_{\mathrm{LP}}^{(\ell)}, \\
F_{\mathrm{HP}}^{(\ell)} &= (I - \hat{A}) H_{\text{spat}}^{(\ell-1)} W_{\mathrm{HP}}^{(\ell)}, \\
g^{(\ell)} &= \tanh\bigl[F_{\mathrm{LP}}^{(\ell)}  \|  F_{\mathrm{HP}}^{(\ell)}\bigr], \\
H_{\text{spat}}^{(\ell)} &= \sigma \left(g^{(\ell)} \odot F_{\mathrm{LP}}^{(\ell)}
    + (1 - g^{(\ell)}) \odot F_{\mathrm{HP}}^{(\ell)}\right).
\end{aligned}
\end{align}
This disentangles low- ($F^{(\ell)}_{\mathrm{LP}}$) and high-pass ($F^{(\ell)}_{\mathrm{HP}}$) signals with learnable gate $g^{(\ell)}$ for spatial filtering. To explicitly encourage heterophily specialization and feature-noise robustness, we also penalize over-smoothing in the spatial branch via a high-pass energy emphasis as follows:
\begin{equation}
\boxed{\mathcal{R}_{\text{HP}} =
-\sum_{\ell=1}^{L_p}
\left\|
  \mathcal{L}^{1/2} H_{\text{spat}}^{(\ell)}
\right\|_F^2}
\end{equation}
which encourages retaining high-frequency components (negative sign denotes encouraging, not penalizing). The final output is $Z_{\rm spat} := H_{\rm spat}^{(L_p)} \in \mathbb{R}^{n\times d_\ell}$.

\subsection{Adaptive Branch Fusion}
\label{sec:fusion}
First, we diagnose how sensitive each branch is to attacks via gradient-based measures:
\begin{align}
\begin{aligned}
\label{eq_branch_robust}
r_{i}^{A,\text{spec}} &= 
\left\|
\frac{\partial \mathcal{L}_{\text{CE}}}{\partial A_{i,:}}
\right\|_1,\qquad
r_{i}^{X,\text{spec}} =
\left\|
\frac{\partial \mathcal{L}_{\text{CE}}}{\partial X_{i,:}}
\right\|_1,\\ 
r_{i}^{A,\text{spat}} &= 
\left\|
\frac{\partial \mathcal{L}_{\text{CE}}}{\partial A_{i,:}}
\right\|_1,\qquad
r_{i}^{X,\text{spat}} =
\left\|
\frac{\partial \mathcal{L}_{\text{CE}}}{\partial X_{i,:}}
\right\|_1,
\end{aligned}
\end{align}
where $\mathcal{L}_{\text{CE}}$ is computed via Eq. \ref{eq:cross} and updated separately when back-propagating through each branch alone. Intuitively, if $r_{i}^{\text{spec}}$ is high, the spectral branch is fragile to structural changes, so fusion should rely less on it for edge-attacks, and vice versa.

\paragraph{Context-aware gating.}
Remember that $Z_{\text{spec}}$ and $Z_{\text{spat}}$ are the node embeddings from each branch. We define node-wise context to guide fusion as follows:
\begin{equation}
\alpha  =  \sigma \Bigl(\mathrm{MLP}_{\varphi}\bigl(
[ Z_{\text{spec}} \| Z_{\text{spat}} \| r^A \| r^X ]
\bigr)\Bigr)
\in (0,1)^{n\times d_\ell},
\label{eq:fusion_gate}
\end{equation}
where $r^A, r^X$ collect robustness signals for each node and channel (Eq. \ref{eq_branch_robust}) since we broadcast or summarize gradients channel-wise. The fused representation is given by:
\begin{equation}
  Z  =  \alpha \odot Z_{\text{spec}}  +  (1-\alpha) \odot Z_{\text{spat}}.
  \label{eq_merge}
\end{equation}
This gate learns to emphasize the spectral branch when the neighborhood is homophilic and resilient to feature attacks (or fragile to structure attacks), and vice versa for the spatial branch. A linear classifier on $Z$ yields the logits and the standard cross-entropy loss as below:
\begin{equation}
  \boxed{\mathcal{L}_{\text{CE}}(\theta;A,X,Y)
   = 
  \mathcal{L}_{\text{nll}}
  \bigl(\text{softmax}(Z) , Y\bigr)}
  \label{eq:cross}
\end{equation}

\subsection{Branch Specialized Adversarial Training}
\label{sec:robust}
The adversaries can (i) flip up to $p$ edges ($\ell_0$ budget) and (ii) perturb node features within $\|\Delta X\|_\infty \le \varepsilon$ ($\ell_\infty$ budget):
\begin{equation}
\mathcal{S}(G;p,\varepsilon)  = 
\bigl\{(\mathbf{A}',\mathbf{X}'): 
\|\mathbf{A}'-\mathbf{A}\|_{0} \le p, 
\|\mathbf{X}'-\mathbf{X}\|_{\infty}\le\varepsilon
\bigr\}.
\label{threat_model}
\end{equation}

\paragraph{Branch-specific inner maximization.}
To specialize in robustness, we decouple adversarial supervision:

(i) Spectral branch (edge perturbation)
\begin{equation}
\boxed{\mathcal{L}_{\text{adv}}^{A} = \max_{\|\mathbf{A}'-\mathbf{A}\|_{0}\le p}
  \mathcal{L}_{\text{CE}} 
  \bigl(f_{\text{spec}}(\mathbf{A}',\mathbf{X});Y\bigr)}
\end{equation}

(ii) Spatial branch (feature perturbation)
\begin{equation}
\boxed{\mathcal{L}_{\text{adv}}^{X} = \max_{\|\mathbf{X}'-\mathbf{X}\|_{\infty}\le\varepsilon}
  \mathcal{L}_{\text{CE}} 
  \bigl(f_{\text{spat}}(\mathbf{A},\mathbf{X}');Y\bigr)}
\end{equation}

(iii) To close the worst‑case gap, we further include  
\begin{equation}
\boxed{
\mathcal L_{\text{adv}}^{A+X}
= \max_{\substack{A',X'}}
  \mathcal L_{\text{CE}} 
  \bigl(f_{\text{fuse}}(A+\Delta A,X+\Delta X);Y\bigr)}
\end{equation}
executed once every $T$ rounds (default $T{=}10$). They are approximated via Projected Gradient Descent (PGD), alternating with SGD updates of the corresponding branch.

\subsection{Full Objective with Conditional Consistency}
\paragraph{Conditional consistency.} Besides the cross entropy and adversarial terms, we use a consistency loss to align branches on unlabeled nodes $\mathcal{V}_U := \mathcal{V} \setminus \mathcal{V}_L$ only when the node is in a stable context (low attack sensitivity and moderate homophily). Let $b_u = \sigma\left(
  \mathrm{MLP}_g([r^A_u  \| r^X_u])
\right)$ be a binary mask that activates consistency based on the branch robustness (Eq. \ref{eq_branch_robust}) as follows:
\begin{equation}
\label{eq_cons}
\boxed{\mathcal{L}_{\text{cons}} =
\sum_{u\in\mathcal{V}_U} 
b_u \cdot
\bigl\|Z_{\text{spec},u} - Z_{\text{spat},u}\bigr\|_2^2}
\end{equation}
Conversely, when $b_u=0$ (high heterophily), we encourage complementarity via a margin-based separation:
\begin{equation}
\label{eq_comp}
\boxed{\mathcal{L}_{\text{comp}} =
\sum_{u\in\mathcal{V}_U} 
(1-b_u)\cdot
\max\bigl\{0, \gamma - \bigl\|Z_{\text{spec},u} - Z_{\text{spat},u}\bigr\|_2\bigr\}^2}
\end{equation}
Thus, the branches do not collapse to identical representations where diversity is beneficial. 

\paragraph{Full objective.} Using hyperparameters $\lambda_{\{\cdot\}} \geq 0$, the overall training objective is given by:
\begin{align}
\label{full_obj}
\min_{\theta} \quad
&\underbrace{\mathcal{L}_{\text{CE}}(\theta;A,X,Y)}_{\text{cross entropy}}
 + 
\lambda_{\mathrm{adv}}(\underbrace{\mathcal{L}_{\text{adv}}^{A}
 + \mathcal{L}_{\text{adv}}^{X} + \mathcal{L}_{\text{adv}}^{A+X}}_{\text{adversarial}})
\nonumber\\
& + 
\lambda_{\mathrm{cons}}(
\underbrace{\mathcal{R}_{\text{LP}} +  \mathcal{R}_{\text{HP}} + 
\mathcal{L}_{\text{cons}} + 
\mathcal{L}_{\text{comp}}}_{\text{branch specialization}}).
\end{align}
Algorithmic details and complexity are in \textbf{Appendix B}.

\section{Theoretical Analysis} \label{sec:theory}
We now rigorously validate the SpecSphere architecture. Our theoretical analysis consists of—\textit{expressivity of gating beyond 1-WL, frequency bias, robustness, and consistency vs complementarity}—with detailed proofs.

\subsection{Expressivity of Node Channel Gating}
\label{sec:expressivity}
We prove how the node channel gate $\alpha$ in Eq. \ref{eq:fusion_gate} enlarges the hypothesis class beyond a simple scalar convex mixture.

\begin{manualdefinition}[Branch Hypothesis Classes]
\label{def:hyp_class}
Let 
\begin{align}
\nonumber
&\mathcal{F}_{\mathrm{spec}}
:= 
\big\{  Z_{\mathrm{spec}} : \exists \theta_{\mathrm{spec}} \big\},   
\mathcal{F}_{\mathrm{spat}}
:=
\big\{  Z_{\mathrm{spat}} : \exists \theta_{\mathrm{spat}}\big\} \\ 
&\mathcal{F}_{\mathrm{mix}}
:=
\Big\{  Z = \alpha \odot Z_{\mathrm{spec}} + (1-\alpha)\odot Z_{\mathrm{spat}} \Big\},
\end{align}
where all maps are permutation–equivariant (Def. \ref{def:eq}).
\end{manualdefinition}

\begin{manuallemma}[Mask Approximation under Continuity]
\label{lm:elem_gate}
Let $\mathcal{D} := [Z_{\mathrm{spec}}\| Z_{\mathrm{spat}}\| r^A\| r^X]$ be a compact set of inputs in Eq. \ref{eq:fusion_gate}. For any continuous function $M:\mathcal{D}\to[0,1]^{n\times d_\ell}$, there exist parameters of $\mathrm{MLP}_\varphi$ such that $\sup_{x\in\mathcal{D}}\|\alpha(x)-M(x)\|_\infty < \varepsilon$.
Consequently,
\begin{equation}
Z(x)=\alpha(x)\odot Z_{\mathrm{spec}}(x)
     +(1-\alpha(x))\odot Z_{\mathrm{spat}}(x)
\end{equation}
can approximate 
$M(x)\odot Z_{\mathrm{spec}}(x) + (1-M(x))\odot Z_{\mathrm{spat}}(x)$
uniformly on $\mathcal{D}$.
\end{manuallemma}

\begin{manualtheorem}[Strict Enlargement over Scalar Convex Mixtures]
\label{thm:strict_superset}
Let 
$\operatorname{conv}(\mathcal{F}_{\mathrm{spec}}\cup \mathcal{F}_{\mathrm{spat}})$
be the set of global convex combinations 
$t Z_{\mathrm{spec}} + (1-t) Z_{\mathrm{spat}}$ with $t\in[0,1]$.
Then,
\begin{equation}
\operatorname{conv}(\mathcal{F}_{\mathrm{spec}}\cup \mathcal{F}_{\mathrm{spat}})
\subsetneq
\mathcal{F}_{\mathrm{mix}},
\end{equation}
meaning that the representational power of node channel gating is stronger than the scalar-based one.

\noindent
\textbf{Proof.} see Appendix C
\end{manualtheorem}
In both cases, $\mathcal{F}_{\mathrm{mix}}$ realizes a mixture-of-experts at the node–channel level, and the number of linear regions (or smooth regimes) grows with the width/depth of $\mathrm{MLP}_\varphi$.

\subsection{Beyond 1-Weisfeiler-Lehman}
We illustrate that SpecSphere exceeds the expressive power of the 1-Weisfeiler-Lehman (1-WL) test, a common theoretical limit for many graph neural networks. Given a graph $G=(V,E)$, node features $X\in\mathbb{R}^{n\times d}$ are informative if there exists at least one feature dimension whose value is not constant on every orbit induced by the graph automorphism group. Equivalently, $X$ is not a constant multiset over any 1-WL indistinguishable pair of nodes.

\begin{manualtheorem}[Beyond 1-WL via Node Channel Gating]\label{thm:strict}
Let $G$ and $G'$ be the 10-vertex Cai-Fürer-Immerman (CFI) graphs \cite{cai1992optimal}, which are (i) indistinguishable by the 1-WL test (Def. \ref{def:1wl}) and (ii) co-spectral for the normalized Laplacian. Assume node features $X$ are informative. Consider SpecSphere with:
(i) a spectral branch using a Chebyshev filter of order $K\ge 1$, (ii) a spatial branch with attention-based aggregation, and (iii) a node--channel gate $\alpha=\sigma \big(\mathrm{MLP}_\varphi([Z_{\mathrm{spec}}\|Z_{\mathrm{spat}}\|r^A\|r^X])\big)$. Then, there exist Chebyshev coefficients $\{\alpha_k\}_{k=0}^{K}$, spatial parameters $\theta_{\mathrm{spat}}$, and gating weights $\theta_\varphi$ such that
\begin{equation}
\|Z(G)-Z(G')\|_\infty > 0,
\end{equation}

\noindent
\textbf{Proof}. see Appendix D
\end{manualtheorem}
The above theorem implies that SpecSphere distinguishes $G$ and $G'$, going beyond the power of 1-WL. Conversely, suppose all nodes share identical features and the gate degenerates to a global scalar (or any multiset-invariant function of messages). In that case, any message-passing GNN (including SpecSphere under this degeneration) maps $G$ and $G'$ to identical embeddings at every layer.

\subsection{Spectral-Spatial Frequency Bias}
\label{sec:freq_bias}
We aim to show how SpecSphere learns to specialize each branch for different frequency regimes in the graph spectrum. Let $\mathcal{L} = U \Lambda U^\top$ be the normalized Laplacian eigen-decomposition, where $U = [u_1,\dots,u_n]$ is orthonormal. For any layer output $H$, we can induce
\begin{equation}
\|\mathcal{L}^{1/2} H\|_F^2
= \sum_{i=1}^n \lambda_i   \|u_i^\top H\|_F^2
= \sum_{i=2}^n \lambda_i   \|u_i^\top H\|_F^2,
\end{equation}
since $\lambda_1=0$. Thus, the regularizers $\mathcal{R}_{\text{LP}}$ and $\mathcal{R}_{\text{HP}}$ can be written as weighted sums of spectral coefficients below.

\begin{manuallemma}[$\mathcal{R}_{\text{LP}}$ Bound]
\label{lem:energy_decomp}
Fix $\lambda_{\text{cons}}>0$ and consider minimizing $\mathcal{J}_{\text{spec}} = \mathcal{L}_{\text{CE}} + \lambda_{\text{cons}}\mathcal{R}_{\text{LP}}$ for $\theta_{\text{spec}}$. Let $C_{\text{spec}}$ be an upper bound on $\mathcal{L}_{\text{CE}}$ over feasible parameters. Then,
\begin{equation}
\label{eq_low}
\sum_{\ell=1}^{L_s}\sum_{i=2}^n \lambda_i  \|u_i^\top H_{\text{spec}}^{(\ell)}\|_F^2
 \le  \frac{C_{\text{spec}}}{\lambda_{\text{cons}}}.
\end{equation}
Therefore, high-frequency components (large $\lambda_i$) are uniformly suppressed, yielding a low-pass bias. 
\end{manuallemma}

\begin{manuallemma}[$\mathcal{R}_{\text{HP}}$ Bound]
For the spatial branch, the objective contains $-\lambda_{\text{cons}}\sum_{\ell}\|\mathcal{L}^{1/2}H_{\text{spat}}^{(\ell)}\|_F^2$. Equivalently, define $\mathcal{J}_{\text{spat}}
= \mathcal{L}_{\text{CE}} - \lambda_{\text{cons}} \mathcal{R}_{\text{HP}}$.
Minimizing $\mathcal{J}_{\text{spat}}$ encourages maximizing the Laplacian energy term subject to performance constraints, i.e.,
\begin{equation}
\label{eq_high}
\sum_{\ell,i}\lambda_i \|u_i^\top H_{\text{spat}}^{(\ell)}\|_F^2
 \ge  \frac{1}{\lambda_{\text{cons}}}\bigl(\widetilde{C}_{\text{spat}} - \mathcal{L}_{\text{CE}}\bigr),
\end{equation}
for some constant $\widetilde{C}_{\text{spat}}$ depending on model capacity. Thus, high-frequency (heterophilic) components are retained or amplified, mitigating over-smoothing.
\end{manuallemma}

\begin{manualcorollary}[Two-sided Frequency Separation]
\label{cor:two_sided}
Combining Eq. \ref{eq_low} and \ref{eq_high}, the learned representations satisfy
\begin{align}
\begin{aligned}
\sum_{i>\kappa}\lambda_i \|u_i^\top H_{\text{spec}}\|_F^2 \ll 
\sum_{i\le \kappa}\lambda_i \|u_i^\top H_{\text{spec}}\|_F^2,
\\
\sum_{i\le \kappa}\lambda_i \|u_i^\top H_{\text{spat}}\|_F^2 \ll 
\sum_{i>\kappa}\lambda_i \|u_i^\top H_{\text{spat}}\|_F^2,
\end{aligned}
\end{align}
for some cut index $\kappa$ determined by optimization. Therefore, the spectral branch dominates lower frequencies, while the spatial branch dominates higher one.
\end{manualcorollary}

\begin{manualtheorem}[Homophily/Heterophily Adaptation via Frequency Bias]
\label{thm:homo_bias}
Assume labels are predominantly encoded in low-frequency components when local homophily is high, and otherwise in high-frequency components. Under the objectives with $\mathcal{R}_{\text{LP}}$ and $\mathcal{R}_{\text{HP}}$, there exists a gate $\alpha$ such that the fused output $Z$ attains (up to $\varepsilon$) the lower of the two Bayes risks associated with each frequency band. 

\noindent
\textbf{Proof.} see Appendix E
\end{manualtheorem}

\subsection{Certified Robustness}\label{sec:robustness}
We certify robustness under the threat model $(A+\Delta A,X+\Delta X) \in \mathcal{S}(G;p,\varepsilon)$ in Eq. \ref{threat_model}, and show that each branch is robust against different types of perturbations. 

\paragraph{Branch-wise Lipschitz bounds.}
Rescale $\tilde{\mathcal L}\in[-1,1]$. For Chebyshev filters, 
$\|T_k(\tilde{\mathcal L})\|_2\le 1$ and 
$\|T_k(\tilde{\mathcal L}+\Delta)-T_k(\tilde{\mathcal L})\|_2 \le k2^{k-1}\|\Delta\|_2$ with coefficients $\mathbf c$. Then,
\begin{align}
\label{eq:spec_bound}
\|Z_{\mathrm{spec}}(A+\Delta A)-Z_{\mathrm{spec}}(A)\|_\infty
&\le B_{\mathrm{spec}}^{A} \|\Delta A\|_2, 
\\ \nonumber
\|Z_{\mathrm{spec}}(X+\Delta X)-Z_{\mathrm{spec}}(X)\|_\infty
&\le B_{\mathrm{spec}}^{X} \|\Delta X\|_\infty,
\end{align}
with 
$B_{\mathrm{spec}}^{A}:=(2^{K+1}-1)\|\mathbf c\|_1\|X\|_\infty, B_{\mathrm{spec}}^{X}:=(2^{K+1}-1)\|\mathbf c\|_1$. Assume each attention layer and weight spectral norm $\le\beta$ satisfies $\big\|\partial\hat A/\partial A\big\|_2\le 1$ in spatial branch. Then,
\begin{align}
\label{eq:spat_bound}
\|Z_{\mathrm{spat}}(A+\Delta A)-Z_{\mathrm{spat}}(A)\|_\infty
&\le B_{\mathrm{spat}}^{A} \|\Delta A\|_2,
\\ \nonumber
\|Z_{\mathrm{spat}}(X+\Delta X)-Z_{\mathrm{spat}}(X)\|_\infty
&\le B_{\mathrm{spat}}^{X} \|\Delta X\|_\infty,
\end{align}
where $B_{\mathrm{spat}}^{A}:=\beta^{L_p}\|X\|_\infty,  B_{\mathrm{spat}}^{X}:=\beta^{L_p}$.

\paragraph{Gate Lipschitz.}
Let $\alpha=\sigma(\mathrm{MLP}_\varphi([\cdot]))$ with weight matrices satisfying $\|W_\ell\|_2\le\gamma_\ell$, where we define $L_\varphi:=\prod_{\ell=1}^{L_f}\gamma_\ell$. Since $\sigma'(z)\le 1/4$, the gate is $L_{\text{gate}}$‑Lipschitz with
\begin{align}
\begin{aligned}
L_{\text{gate}}  :=  \frac{1}{4}L_\varphi\Big(B_{\mathrm{spec}}^{A}+B_{\mathrm{spat}}^{A}\Big) \quad\text{(w.r.t.\ $A$)},\\
\tilde L_{\text{gate}}  :=  \frac{1}{4}L_\varphi\Big(B_{\mathrm{spec}}^{X}+B_{\mathrm{spat}}^{X}\Big) \quad\text{(w.r.t.\ $X$)}.
\end{aligned}
\end{align}

\begin{manualtheorem}[Certified Bound for Branch-Specialized Fusion]
\label{thm:branch_cert}
Let $Z=\alpha\odot Z_{\mathrm{spec}} + (1-\alpha)\odot Z_{\mathrm{spat}}$. For any $\mathcal{S}(G;p,\varepsilon)$ with $\|\Delta A\|_2\le \sqrt{2p}$ and $\|\Delta X\|_\infty\le \varepsilon$, we have
\begin{align}
\begin{aligned}
&\|Z(A+\Delta A,X+\Delta X)-Z(A,X)\|_\infty
\le \\ &\big(1+L_{\text{gate}}\big)B^A\sqrt{2p} + \big(1+\tilde L_{\text{gate}}\big)B^X\varepsilon,
\label{eq:branch_cert_bound}
\end{aligned}
\end{align}
where $B^A := B_{\mathrm{spec}}^{A}+B_{\mathrm{spat}}^{A}$ and $B^X := B_{\mathrm{spec}}^{X}+B_{\mathrm{spat}}^{X}$.

\noindent
\textbf{Proof.} See Appendix F
\end{manualtheorem}
Consequently, if the right-hand side of Eq. \ref{eq:branch_cert_bound} is strictly smaller than the classification margin $\gamma$, the predicted labels are invariant to any $(p,\varepsilon)$‑budget attacks.

\begin{table*}[!t]
\centering
\small
\caption{Node classification accuracy on benchmark datasets under robustness attacks. Bold indicates the best performance in each column. Symbol $+\Delta$ denotes that the adversarial training was applied during training (see Eq. \ref{full_obj})}
\label{results}
\begin{adjustbox}{width=\textwidth}
\begin{tabular}{lcccccccc}
\toprule
 & \multicolumn{4}{c}{Cora (\textit{homophilous})} & \multicolumn{4}{c}{Chameleon (\textit{heterophilous})} \\
\cmidrule(r){2-5} \cmidrule(l){6-9}
Perturbation Type
  & Clean & DropEdge & Metattack & PGD
  & Clean & DropEdge & Metattack & PGD\\
\textit{Attack ratio} & x & 20\% & 5\% & $\varepsilon{=}0.1$ & x & 20\% & 5\% & $\varepsilon{=}0.1$ \\
\midrule
$[1]$ GCN (spectral)          
  & 81.5$_{\pm0.4}$ & 69.5$_{\pm2.3}$ & 56.3$_{\pm1.1}$ & 72.8$_{\pm0.8}$
  & 45.9$_{\pm1.7}$ & 32.5$_{\pm0.9}$ & 29.6$_{\pm2.5}$ & 28.7$_{\pm1.0}$\\
    + $\Delta$              
  & 80.1$_{\pm1.2}$ & 78.5$_{\pm0.6}$ & 70.1$_{\pm1.4}$ & 79.3$_{\pm1.5}$
  & 44.2$_{\pm0.8}$ & 37.2$_{\pm1.6}$ & 34.4$_{\pm0.7}$ & 36.0$_{\pm1.2}$\\
$[2]$ GAT (spatial)           
  & 83.0$_{\pm0.5}$ & 70.9$_{\pm1.2}$ & 58.2$_{\pm2.6}$ & 74.2$_{\pm1.0}$
  & 47.4$_{\pm1.4}$ & 31.8$_{\pm0.7}$ & 28.8$_{\pm1.9}$ & 29.7$_{\pm1.3}$\\
    + $\Delta$              
  & 81.8$_{\pm0.9}$ & 79.7$_{\pm1.8}$ & 71.6$_{\pm0.5}$ & 80.0$_{\pm0.7}$
  & 46.2$_{\pm1.1}$ & 38.3$_{\pm0.4}$ & 34.1$_{\pm1.2}$ & 37.5$_{\pm2.0}$\\
$[1$+$2]$ \textbf{SpecSphere}                
  & \textbf{83.4}$_{\pm0.4}$ & 71.8$_{\pm1.6}$ & 60.0$_{\pm0.8}$ & 75.8$_{\pm1.3}$
  & \textbf{48.8}$_{\pm1.2}$ & 34.4$_{\pm1.1}$ & 31.5$_{\pm1.0}$ & 31.8$_{\pm1.5}$\\
    + $\Delta$              
  & 83.0$_{\pm0.7}$ & \textbf{81.4}$_{\pm0.7}$ & \textbf{75.0}$_{\pm1.3}$ & \textbf{81.5}$_{\pm0.6}$
  & 48.2$_{\pm0.6}$ & \textbf{43.7}$_{\pm2.0}$ & \textbf{39.1}$_{\pm0.9}$ & \textbf{42.5}$_{\pm1.1}$\\
\midrule
$[3]$ APPNP (spectral)        
  & 83.4$_{\pm0.8}$ & 68.9$_{\pm1.5}$ & 53.4$_{\pm0.6}$ & 70.1$_{\pm1.4}$
  & 45.5$_{\pm1.0}$ & 40.0$_{\pm1.8}$ & 35.2$_{\pm1.1}$ & 38.1$_{\pm0.9}$\\
    + $\Delta$              
  & 82.0$_{\pm0.6}$ & 75.3$_{\pm2.1}$ & 67.2$_{\pm1.3}$ & 77.5$_{\pm0.8}$
  & 44.9$_{\pm1.2}$ & 39.3$_{\pm0.5}$ & 35.5$_{\pm2.0}$ & 41.1$_{\pm1.4}$\\
$[4]$ FAGCN (spatial)         
  & 82.8$_{\pm0.4}$ & 70.4$_{\pm2.0}$ & 55.1$_{\pm1.0}$ & 73.0$_{\pm1.2}$
  & 47.3$_{\pm1.6}$ & 41.1$_{\pm1.2}$ & 36.7$_{\pm0.8}$ & 40.2$_{\pm1.5}$\\
    + $\Delta$              
  & 82.1$_{\pm1.0}$ & 77.6$_{\pm0.9}$ & 69.0$_{\pm1.4}$ & 78.4$_{\pm0.6}$
  & 46.6$_{\pm0.7}$ & 40.9$_{\pm2.1}$ & 37.2$_{\pm1.3}$ & 42.6$_{\pm1.1}$\\
$[3$+$4]$ \textbf{SpecSphere}                
  & \textbf{83.9}$_{\pm0.5}$ & 74.1$_{\pm1.8}$ & 60.4$_{\pm0.6}$ & 75.7$_{\pm1.2}$
  & \textbf{47.9}$_{\pm1.3}$ & 42.3$_{\pm1.2}$ & 39.1$_{\pm1.7}$ & 43.5$_{\pm0.9}$\\
    + $\Delta$              
  & 83.1$_{\pm0.6}$ & \textbf{80.0}$_{\pm0.8}$ & \textbf{74.2}$_{\pm1.2}$ & \textbf{80.1}$_{\pm0.4}$
  & 47.0$_{\pm1.1}$ & \textbf{45.4}$_{\pm1.7}$ & \textbf{43.4}$_{\pm1.0}$ & \textbf{45.0}$_{\pm0.9}$\\
\bottomrule
\end{tabular}
\end{adjustbox}
\end{table*}

\subsection{Consistency–Complementarity Trade-off}
\label{sec:cons_comp}
We formalize how SpecSphere achieves a principled balance between consistency and complementarity across spectral and spatial representations. Recall the soft mask $b_u=\sigma(\mathrm{MLP}_g([r^A_u\|r^X_u]))\in(0,1)$ in Eq. \ref{eq_cons} and \ref{eq_comp} for conditional consistency.

\begin{manualdefinition}[Masked Discrepancy and Margin Set]\label{def:sets}
Let $\Delta_u := \|Z_{\mathrm{spec},u}-Z_{\mathrm{spat},u}\|_2$.
Define the consistency set $\mathcal{C}:=\{u\in\mathcal V_U: b_u\ge \tfrac12\}$ and the complementarity set $\mathcal{D}:=\mathcal V_U\setminus\mathcal{C}$.
\end{manualdefinition}

\begin{manuallemma}[Upper Bound on Consistency Discrepancy]\label{lem:cons_bound}
Let $B:=\sum_{u\in\mathcal V_U} b_u>0$. Then,
\begin{equation}
\frac{1}{B}\sum_{u\in\mathcal V_U} b_u \Delta_u^2 \le 
\frac{\mathcal{L}_{\text{cons}}^\star}{B},
\end{equation}
where $\mathcal{L}_{\text{cons}}^\star$ is the value of $\mathcal{L}_{\text{cons}}$ at a (local) minimum. Thus, on nodes with large $b_u$, the average branch discrepancy is tightly controlled.
\end{manuallemma}

\begin{manuallemma}[Lower Separation via Margin Hinge]\label{lem:comp_margin}
For any $u\in\mathcal{V}_U$, the complementarity term contributes $(1-b_u)(\gamma-\Delta_u)^2 \leq \mathcal{L}_{\text{comp}}^\star$ if $\Delta_u < \gamma$. Thus, the following inequality is satisfied when $b_u < 1$,
\begin{equation}
\Delta_u
 \ge 
\gamma - \sqrt{\frac{\mathcal{L}_{\text{comp}}^\star}{1-b_u}}.
\end{equation}
In particular, for $u\in\mathcal{D}$ with $b_u\le\tfrac12$, we can obtain the inequality $\Delta_u \ge \gamma - \sqrt{2 \mathcal{L}_{\text{comp}}^\star}$.
\end{manuallemma}
Suppose $\lambda_{\text{c}}>0$ in Eq. \ref{full_obj}. 
At any stationary point of the total loss, there is no trivial collapse ($Z_{\mathrm{spec}}=Z_{\mathrm{spat}}$) on $\mathcal{D}$ unless $\mathcal{L}_{\text{comp}}^\star=0$ and $\gamma=0$. Conversely, $\mathcal{L}_{\text{cons}}$ forces $\Delta_u$ to be small in expectation on $\mathcal{C}$. Therefore, the model simultaneously achieves agreement on stable nodes and diversity on unstable (or heterophilic) nodes.

\begin{manualtheorem}[Optimal Trade-off via Learnable Mask]\label{thm:tradeoff}
Assume $\mathrm{MLP}_g$ is a universal approximator on a compact domain of robustness signals. Then, there exist parameters of $\mathrm{MLP}_g$ such that $b_u \approx \tilde b_u$ uniformly for any target weighting scheme $\tilde b_u\in(0,1)$ that minimizes the ideal risk below:
\begin{equation}
\mathcal{R}_{\text{ideal}}
= \sum_{u\in\mathcal V_U}\Big[ 
\tilde b_u \Delta_u^2
+ (1-\tilde b_u) \max\{0,\gamma-\Delta_u\}^2
\Big],
\end{equation}
Consequently, SpecSphere can approximate the optimal balance between consistency and complementarity, achieving the minimum of $\mathcal{R}_{\text{ideal}}$ up to arbitrary precision.

\noindent
\textbf{Proof}. see Appendix G
\end{manualtheorem}

\section{Experiment}
We conduct experiments on benchmark datasets, including robustness analyses, ablation studies, and parameter‑sensitivity experiments. Due to space constraints, detailed descriptions of the datasets, we introduce more results and statistics of the node classification results, and robust GNNs in \textbf{Appendix H}.

\subsection{Baselines and Implementation} \label{subsec:baselines}

\textbf{Baselines.} As illustrated in Table \ref{results}, we include two spectral methods: GCN \cite{kipf2016semi} and APPNP \cite{klicpera2018predict}; and two spatial methods: GAT \cite{velickovic2017graph} and FAGCN \cite{bo2021beyond} as baselines. We fuse their outputs to form a SpecSphere model for each spectral–spatial pair.

\noindent
\textbf{Implementation.} We implement our model in PyTorch Geometric on a single NVIDIA TITAN Xp GPU (12GB memory). Each model is trained for 300 epochs with early stopping (patience = 100) using the validation split. All datasets use the public Planetoid splits \cite{kipf2016semi}. The full graph is processed each step with learning rate=$10^{-2}$, dropout=0.5, and weight‑decay=$5\times10^{-4}$. For the spatial branch, we use 8 attention heads and LeakyReLU. The branches have $L_s=L_p=2$ layers, each with 32 hidden dimensions. In adversarial training, we set $p=\lfloor0.1|\text{E}|\rfloor$ for edge flips and $\varepsilon=0.1$ for feature attacks. PGD uses step‑size=0.01 per step for topology after each update.

\subsection{Main Result}
In Table \ref{results}, SpecSphere achieves the best clean accuracy and the strongest robustness across all perturbations on both Cora (homophilous) and Chameleon (heterophilous). On Cora, the vanilla model attains the highest clean accuracy, outperforming the best single‐branch baseline, and remains the top performer under DropEdge, Metattack, and Feature‐PGD. After PGD adversarial training (+$\Delta$), our model narrows its clean-to-attack gap to under 1\%, exceeding the next best model on each adversarial attack. Averaged over all four conditions, SpecSphere+$\Delta$ scores 79.6\% in comparison to 77.0\% for the runner-up. On Chameleon, similarly leads in clean accuracy and maintains the largest margins under all three attacks. With adversarial training, it outperforms every competitor by 1.2–3.7\% on each perturbation. These results confirm that fusing spectral and spatial branches simultaneously enhances clean accuracy and delivers consistently superior, dataset-agnostic defenses against both random and adversarial graph perturbations.

\begin{figure}
    \centering
    \includegraphics[width=\linewidth]{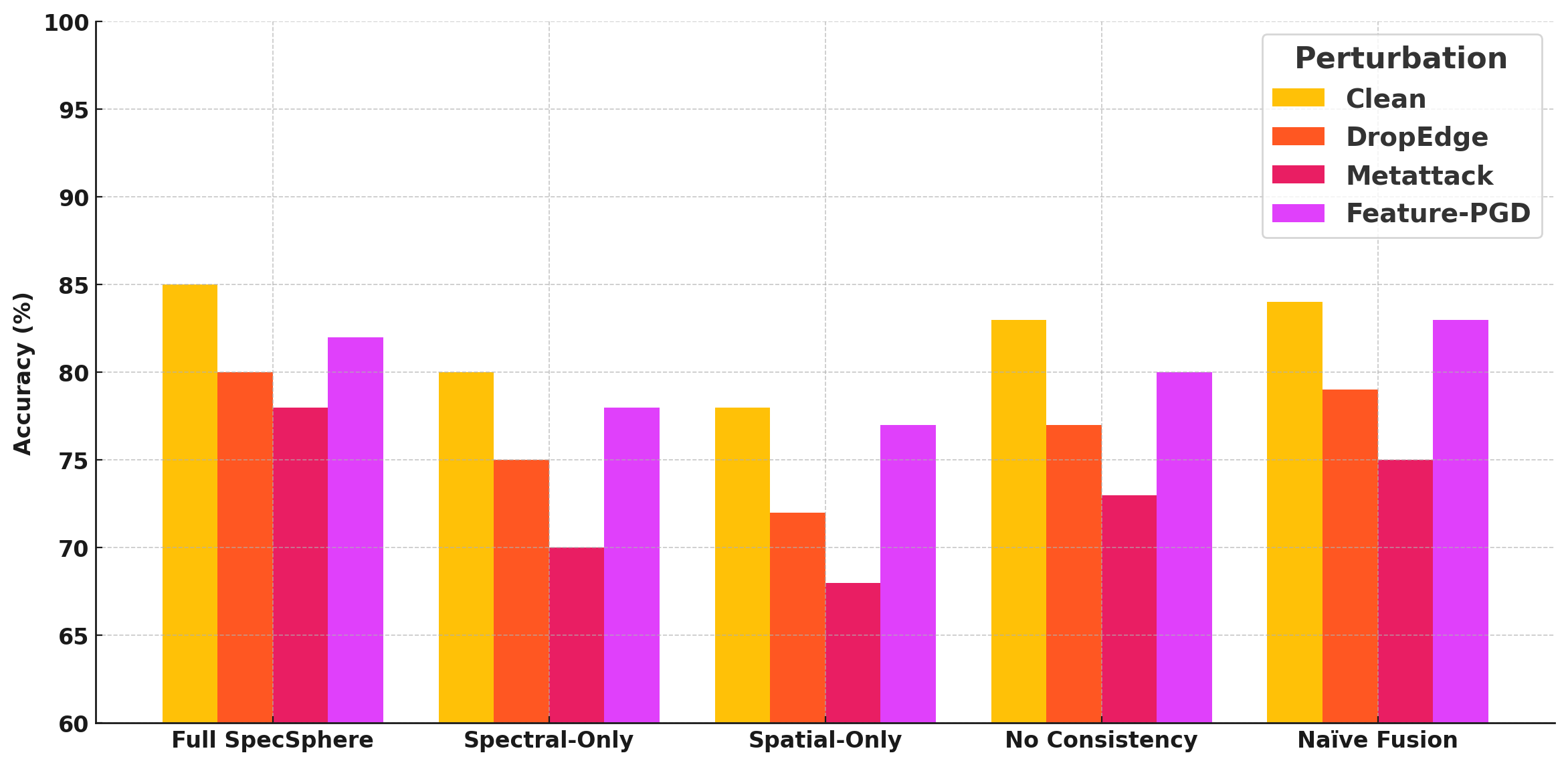}
    \caption{As illustrated in the box, we describe node classification accuracy (y-axis) on Cora under four perturbation regimes (Clean, DropEdge, Metattack, Feature-PGD), comparing five SpecSphere variants (x-axis)}
    \label{ablation_study}
\end{figure}

\subsection{Ablation Study}
\label{sec:ablation}
Figure \ref{ablation_study} compares five variants of SpecSphere on Cora under four perturbation regimes: Clean, DropEdge, Metattack, and Feature-PGD. The \textbf{Full SpecSphere} uses spectral and spatial branches with robustness and consistency regularization. The others include: \textbf{Spectral-Only} removes the spatial branch entirely and trains using only the spectral-branch output. \textbf{Spatial-Only} uses the spatial (message passing) branch only. \textbf{No consistency} retains both branches and the learned fusion, but disables the consistency regularization loss during training to measure the impact of forcing branch outputs to agree. \textbf{Naïve Fusion} uses both branches and consistency regularization, but replaces the learned fusion strategy with a simple average of the two branch outputs. The Clean model yields the highest clean accuracy and robustness across all attacks. Removing one branch (Spectral or Spatial) consistently degrades accuracy, demonstrating that both information are complementary. Disabling consistency regularization also hurts robustness, indicating that aligning the two branches helps the model resist perturbations. Lastly, replacing the learned fusion with a simple average (Naïve Fusion) yields intermediate performance, suggesting that the adaptive fusion strategy provides an additional gain. Overall, this study reinforces that combining dual branches, the specific fusion mechanism, and the consistency loss is critical to achieving strong accuracy and adversarial resilience.

\begin{figure}
    \centering
    \includegraphics[width=\linewidth]{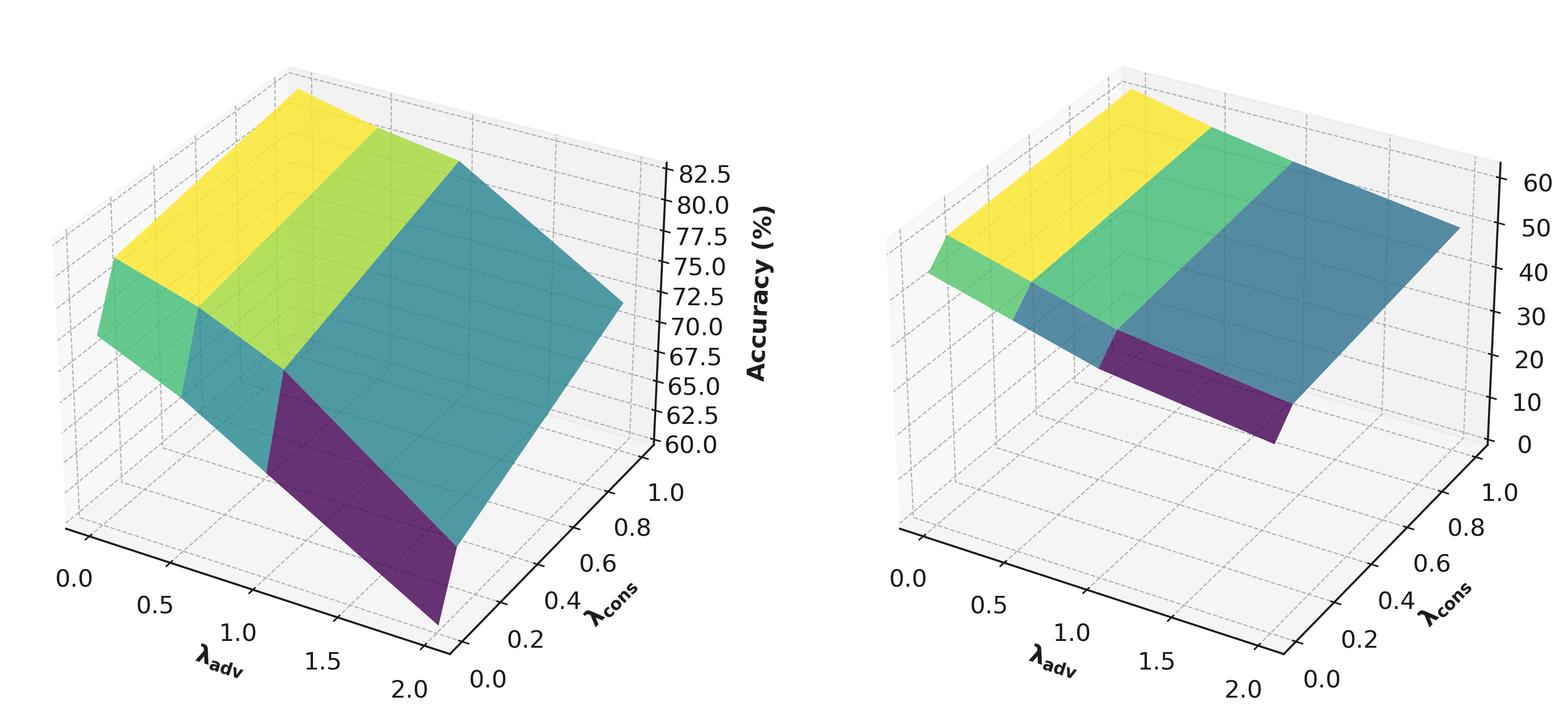}
    \caption{Node classification accuracy of SpecSphere by varying robustness hyperparameters in Eq. \ref{full_obj}, which includes $\lambda_{\text{adv}}$ (x-axis) and $\lambda_{\text{cons}}$ (y-axis). Left and right figure represents the Cora and Chameleon dataset, respectively}
    \label{param_sense}
\end{figure}

\subsection{Hyperparameter Settings}
In Figure \ref{param_sense}, we perform a grid search over the two robustness hyperparameters defined in Eq. \ref{full_obj}, which are adversarial $\lambda_{\mathrm{adv}}\in\{0.0, 0.1, 0.5, 1.0, 2.0\}$ and consistency-complementarity $\lambda_{\mathrm{cons}}\in\{0.0, 10^{-3}, 0.01, 0.1, 1.0\}$ losses. For each $(\lambda_{\mathrm{adv}},\lambda_{\mathrm{cons}})$ pair, we train three times with different random seeds and plot the mean validation accuracy under adversarial edge attack (DropEdge 20\%). All other settings are held constant across datasets. We show 3D surface plots for two representative graphs: Cora (left, homophilic) and Chameleon (right, heterophilic). On Cora, accuracy peaks at $\lambda_{\mathrm{adv}}=0.5$ and $\lambda_{\mathrm{cons}}=10^{-2}$, indicating that moderate adversarial training combined with a small consistency term is ideal. Performance drops when $\lambda_{\mathrm{adv}}\ge1.0$ (overweighting the adversarial loss) or when $\lambda_{\mathrm{cons}}<10^{-3}$ (insufficient branch alignment). On Chameleon, which is heterophilous, the sweet spot shifts to $\lambda_{\mathrm{adv}}=1.0$ and $\lambda_{\mathrm{cons}}=10^{-1}$, suggesting that stronger adversarial regularization and a larger consistency penalty benefit heterophilic graphs. A consistency weight below $10^{-3}$ fails to align the branches adequately, while excessive consistency ($\ge1.0$) impedes the model’s ability to adapt its branch fusion.

\section{Conclusion}
We introduce a dual‑branch graph learner that couples a spectral branch with a spatial branch, each specialized for a different attack type, and integrates node‑wise adaptive gating with branch‑specific adversarial objectives. We prove that the gate learns a channel‑level, non‑linear mixture, strictly enlarging the hypothesis class beyond scalar convex combinations and surpassing the 1‑WL expressiveness barrier. Moreover, the branch‑specialized PGD training scheme yields certified bounds against $(p,\varepsilon)$ edge and feature attacks for the fused model. Empirically, SpecSphere attains state‑of‑the‑art accuracy across multiple benchmarks under both clean and adversarial settings, while retaining the linear‑time scalability of classical Chebyshev‑filtered GNNs. These results demonstrate that the simple design principle of attack‑type specialization with adaptive gating offers a practical path toward robust graph learning on dynamic, partially observed networks.

\section{Ethical Consideration}
As a graph algorithm, our method leverages benchmark graph datasets to improve robustness, which are widely used in the literature and contain public graph structures with anonymized node features and labels. SpecSphere is designed to resist adversarial perturbations, including feature and edge manipulations. This contributes to safety in adversarial settings (e.g., spam detection, fraud analysis). However, any robustness mechanism could also be misused to create resilient misinformation or evasion systems. We urge responsible deployment with appropriate safeguards.

\bibliography{aaai25}

\newpage

\subsection{Reproducibility Checklist}
This paper:
\begin{itemize}
    \item Includes a conceptual outline and/or pseudocode description of AI methods introduced: \textbf{yes}
    \item Clearly delineates statements that are opinions, hypothesis, and speculation from objective facts and results: \textbf{yes}
    \item Provides well marked pedagogical references for less-familiare readers to gain background necessary to replicate the paper: \textbf{yes}
    \item Does this paper make theoretical contributions? \textbf{yes}
    \item All assumptions and restrictions are stated clearly and formally. \textbf{yes}
    \item All novel claims are stated formally (e.g., in theorem statements). \textbf{yes}
    \item Proofs of all novel claims are included. \textbf{yes}
    \item Proof sketches or intuitions are given for complex and/or novel results. \textbf{yes}
    \item Appropriate citations to theoretical tools used are given. \textbf{yes}
    \item All theoretical claims are demonstrated empirically to hold. \textbf{yes}
    \item All experimental code used to eliminate or disprove claims is included. \textbf{yes}
    \item Does this paper rely on one or more datasets? \textbf{yes}
    \item A motivation is given for why the experiments are conducted on the selected datasets: \textbf{yes}
    \item All novel datasets introduced in this paper are included in a data appendix. \textbf{NA}
    \item All novel datasets introduced in this paper will be made publicly available upon publication of the paper with a license that allows free usage for research purposes. \textbf{NA}
    \item All datasets drawn from the existing literature (potentially including authors’ own previously published work) are accompanied by appropriate citations. \textbf{yes}
    \item All datasets drawn from the existing literature (potentially including authors’ own previously published work) are publicly available. \textbf{yes}
    \item All datasets that are not publicly available are described in detail, with explanation why publicly available alternatives are not scientifically satisficing. \textbf{NA}
    \item Does this paper include computational experiments? \textbf{yes}
    \item This paper states the number and range of values tried per (hyper-) parameter during development of the paper, along with the criterion used for selecting the final parameter setting. \textbf{yes}
    \item Any code required for pre-processing data is included in the appendix. \textbf{NA}
    \item All source code required for conducting and analyzing the experiments is included in a code appendix. \textbf{yes}
    \item All source code required for conducting and analyzing the experiments will be made publicly available upon publication of the paper with a license that allows free usage for research purposes. \textbf{yes}
    \item All source code implementing new methods have comments detailing the implementation, with references to the paper where each step comes from \textbf{yes}
    \item If an algorithm depends on randomness, then the method used for setting seeds is described in a way sufficient to allow replication of results. \textbf{Yes}
    \item This paper specifies the computing infrastructure used for running experiments (hardware and software), including GPU/CPU models; amount of memory; operating system; names and versions of relevant software libraries and frameworks. \textbf{yes}
    \item This paper formally describes evaluation metrics used and explains the motivation for choosing these metrics. \textbf{yes}
    \item This paper states the number of algorithm runs used to compute each reported result. \textbf{yes}
    \item Analysis of experiments goes beyond single-dimensional summaries of performance (e.g., average; median) to include measures of variation, confidence, or other distributional information. \textbf{yes}
    \item The significance of any improvement or decrease in performance is judged using appropriate statistical tests (e.g., Wilcoxon signed-rank). \textbf{yes}
    \item This paper lists all final (hyper-)parameters used for each model/algorithm in the paper’s experiments. \textbf{yes}
\end{itemize}

\newpage

\appendix

\onecolumn

\section*{\centering {\LARGE Technical Appendix}}
\addcontentsline{toc}{section}{Appendix}

\vspace{5em}

\begin{tabularx}{\textwidth}{@{}l X r@{}}
\textbf{Appendix A}   & Extension of Table 1                                             & 2 \\\\
\textbf{Appendix B}   & Optimization Strategy and Computational Cost  & 2-4  \\
    & \textbf{B.1} Optimization Strategy                                                   &  \\
    & \textbf{B.2} Computational Cost                                                   &  \\
    & \textbf{B.3} Memory Cost
    &  \\\\
\textbf{Appendix C}   & Proof of Theorem 1                                      & 4 \\\\
\textbf{Appendix D}   & Proof of Theorem 2               & 4-5 \\\\
\textbf{Appendix E}   & Proof of Theorem 3                                   & 5-6 \\\\
\textbf{Appendix F}   & Proof of Theorem 4   & 6 \\\\
\textbf{Appendix G}   & Proof of Theorem 5                                & 7 \\\\
\textbf{Appendix H}  & Datasets and More Experiments & 7-9 \\
    & \textbf{H.1} Datasets &  \\
    & \textbf{H.2} Node Classification &  \\
    & \textbf{H.3} vs Robust GNNs &  \\
    & \textbf{H.4} Analysis on Large Graphs &  \\
\end{tabularx}

\newpage

\begin{table*}[!t]
\centering
\small
\begin{tabular}{|l|
    >{\centering\arraybackslash}p{1.8cm}|
    >{\centering\arraybackslash}p{2.3cm}|
    >{\centering\arraybackslash}p{1.7cm}|
    >{\centering\arraybackslash}p{1.7cm}|
    >{\centering\arraybackslash}p{2.2cm}|}
\hline
\textbf{Model} & \textbf{Architecture} & \textbf{Certification} & \textbf{Adaptivity} & \textbf{Expressivity} & \textbf{Branch Specialization} \\
\hline
Certified GNN \cite{wang2021certified} & Single   & Certified $\ell_0$                    & No       & No         & No \\
\hline
ACM-GNN \cite{luan2022revisiting}       & Hybrid   & No                                     & Adaptive & Standard   & No \\
\hline
MixHop \cite{abu2019mixhop}             & Spatial  & No                                     & Static   & Standard   & No \\
\hline
Geom-GCN \cite{pei2020geom}             & Hybrid   & No                                     & Static   & Standard   & No \\
\hline
H2GNN \cite{zhu2020beyond}              & Hybrid   & No                                     & Adaptive & Standard   & No \\
\hline
GPR-GNN \cite{chien2020adaptive}        & Spectral & No                                     & Adaptive & Standard   & No \\
\hline
GNNGuard \cite{zhang2020gnnguard}       & Hybrid   & No                                     & Adaptive & Standard   & No \\
\hline
EvenNet \cite{lei2022evennet}           & Spectral & Certified $\ell_\infty$                & Static   & Standard   & No \\
\hline
GARNET \cite{deng2022garnet}            & Spectral & Empirical                              & Adaptive & Standard   & No \\
\hline
RUNG \cite{hou2024robust}               & Hybrid   & No                                     & No       & Standard   & No \\
\hline
S$^2$GNN \cite{geisler2024spatio}       & Hybrid   & No                                     & No       & $>1$-WL    & No \\
\hline
TFE-GNN \cite{duan2024unifying}         & Spectral & No                                     & Adaptive & Standard   & No \\
\hline
PCNet \cite{li2024pc}                   & Dual     & No                                     & Adaptive & Standard   & No \\
\hline
UnGSL \cite{han2025uncertainty}         & Hybrid   & No                                     & No       & Standard   & No \\
\hline
SPCNet \cite{li2025simplified}          & Dual     & Partial                                 & Adaptive & Standard   & No \\
\hline
\textbf{SpecSphere (ours)}              & Dual     & Certified ($\ell_0$, $\ell_\infty$)    & Adaptive & $>1$-WL    & Yes \\
\hline
\end{tabular}
\caption{Comprehensive comparison of spectral–spatial and robustness‑aware GNNs. “Architecture” distinguishes single‑pass (Spectral or Spatial), hybrid (one filter type), and dual‑pass (both). “Certification” indicates provable bounds where available. “Branch Specialization” denotes explicit branch‑specific objectives/training (e.g., different threat models or frequency priors).}
\label{tab:expanded_methods}
\end{table*}

\section{A. Extension of Table \ref{recent_methods}}
In Table \ref{tab:expanded_methods}, we extend Table \ref{recent_methods} to present a comprehensive comparison of seventeen representative graph neural network models. The “Architecture” column classifies each model according to whether it employs a single‐pass filter (Spectral or Spatial), a Hybrid plug‐in of one filter type into another backbone, or a true Dual‐pass design that explicitly interleaves both spectral and spatial filtering. The “Certification” column indicates whether a model provides a provable $\ell_{0}$ robustness bound (either via spectral‐gap analysis or randomized smoothing), empirical robustness defenses, or no formal guarantee. The “Heterophily” column records whether each method supports per‐node adaptation to varying degrees of homophily, and the “Expressivity” column captures theoretical power (e.g., comparison to the 1‐WL test) or the absence of such claims.

Across the table, one observes that early certified defenses (e.g., Certified GNN \cite{wang2021certified} and the Randomized‐Smoothing GNN \cite{bojchevski2020efficient}) are limited to single‐pass filtering and do not adapt to heterophily. Subsequent heterophily‐aware architectures (MixHop \cite{abu2019mixhop}, Geom‐GCN \cite{pei2020geom}, GPR‐GNN \cite{chien2020adaptive}, H2GNN \cite{zhu2020beyond}) improve flexibility under static or adaptive heterophily but lack formal robustness certificates. Plug‐in defense modules such as GNNGuard \cite{zhang2020gnnguard} and UnGSL \cite{han2025uncertainty} introduce empirical or uncertainty‐guided perturbation defenses without a unified filtering architecture or certification. RUNG \cite{hou2024robust} further integrates robust aggregation into message passing, yet remains uncertified.

More recent hybrids like S²GNN \cite{geisler2024spatio}, TFE‐GNN \cite{duan2024unifying}, and the dual‐branch PCNet \cite{li2024pc}/SPCNet \cite{li2025simplified} begin to bridge spectral and spatial filtering but either omit robustness guarantees or only offer partial certificates. Finally, our SpecSphere model stands out as the only Dual‐pass architecture combining a closed‐form ($\ell_{0},\ell_\infty$) certificate (via spectral‐gap analysis), per‐node heterophily adaptation through learnable spectral responses and gated spatial attention, and expressivity beyond the 1‐WL test. This table thus clearly situates SpecSphere at the intersection of certified robustness, heterophily adaptation, and theoretical expressivity, highlighting its unique contributions relative to both classical and recent GNN designs.

\section{B. Optimization and Cost}
\subsection{B.1 Optimization Strategy}
As shown in Algorithm \ref{alg:adv_train_full}, we adopt an adversarial optimization strategy that interleaves adversarial perturbation generation with parameter updates in a single end‑to‑end loop. Given a clean graph $(A,X)$, we first initialize perturbed copies $(A',X')$ and perform $T_{\rm PGD}$ projected gradient–ascent steps: at each step, we compute the gradient of the cross‐entropy loss w\.r.t.\ $A'$ and $X'$, take a sign‐based ascent step of size $\alpha_A$ (for edges) or $\alpha_X$ (for features), and then project the updates onto the $\ell_0$–ball of size $p$ and the $\ell_\infty$–ball of radius $\varepsilon$, respectively.

\begin{algorithm}[!t]
\caption{Training Mechanism of SpecSphere}
\label{alg:adv_train_full}
\begin{algorithmic}[1]
\Require Graph $(A,X)$, labels $Y$, unlabeled set $\mathcal V_U$, params $\theta$, budgets $(p,\varepsilon)$
\Ensure Trained parameters $\theta^\ast$
\While{not converged}
  \State \textbf{1. Branch-specific adversarial examples}
  \State $A' \gets A$
  \For{$t=1$ to $T_A$} \Comment{edge PGD for spectral branch}
    \State $\Delta_A \gets \nabla_{A'} \mathcal{L}_{\text{CE}} \big(f_{\text{spec}}(A',X);Y\big)$
    \State $A' \gets \text{Proj}_{\|\cdot\|_0 \le p} \big(A' + \alpha_A \operatorname{sign}(\Delta_A)\big)$
  \EndFor
  \State $X' \gets X$
  \For{$t=1$ to $T_X$} \Comment{feature PGD for spatial branch}
    \State $\Delta_X \gets \nabla_{X'} \mathcal{L}_{\text{CE}} \big(f_{\text{spat}}(A,X');Y\big)$
    \State $X' \gets \text{Proj}_{\|\cdot\|_\infty \le \varepsilon} \big(X' + \alpha_X \operatorname{sign}(\Delta_X)\big)$
  \EndFor

  \State \textbf{2. Forward passes on clean graph}
  \State Compute $\{H_{\text{spec}}^{(\ell)}\}_{\ell=1}^{L_s}$, $Z_{\text{spec}}$ via Eq. \ref{eq:spec-layer-cheb}
  \State Compute $\{H_{\text{spat}}^{(\ell)}\}_{\ell=1}^{L_p}$, $Z_{\text{spat}}$ via Eq. \ref{eq:spatial-layer}

  \State \textbf{3. Robustness signals}
  \State Backprop w.r.t.\ $A,X$ separately for each branch (freezing the other) to get $r^A, r^X$ (L1 norms)

  \State \textbf{4. Fusion gate \& fused embedding}
  \State $\alpha \gets \sigma \big(\mathrm{MLP}_{\varphi}([Z_{\text{spec}}\|Z_{\text{spat}}\|r^A\|r^X])\big)$
  \State $Z \gets \alpha \odot Z_{\text{spec}} + (1-\alpha)\odot Z_{\text{spat}}$

  \State \textbf{5. Consistency / complementarity weights}
  \State $m \gets \sigma \big(\mathrm{MLP}_g([r^A\|r^X])\big)$
  \State $\mathcal{L}_{\text{cons}} \gets \sum_{u\in\mathcal V_U} b_u  \|Z_{\text{spec},u}-Z_{\text{spat},u}\|_2^2$
  \State $\mathcal{L}_{\text{comp}} \gets \sum_{u\in\mathcal V_U} (1-b_u) \max\{0,\gamma-\|Z_{\text{spec},u}-Z_{\text{spat},u}\|_2\}^2$

  \State \textbf{6. Regularizers \& adversarial losses}
  \State $\mathcal{R}_{\text{LP}} \gets \sum_{\ell=1}^{L_s}\|\mathcal{L}^{1/2}H_{\text{spec}}^{(\ell)}\|_F^2,\quad
         \mathcal{R}_{\text{HP}} \gets -\sum_{\ell=1}^{L_p}\|\mathcal{L}^{1/2}H_{\text{spat}}^{(\ell)}\|_F^2$
  \State $\mathcal{L}_{\text{adv}}^{A} \gets \mathcal{L}_{\text{CE}} \big(f_{\text{spec}}(A',X);Y\big),\quad
         \mathcal{L}_{\text{adv}}^{X} \gets \mathcal{L}_{\text{CE}} \big(f_{\text{spat}}(A,X');Y\big),\quad \mathcal L_{\text{adv}}^{A+X} \gets
  \mathcal L_{\text{CE}} 
  \bigl(f_{\text{fuse}}(A',X');Y\bigr)$
  \State $\mathcal{L}_{\text{CE}} \gets \mathcal{L}_{\text{nll}}(\text{softmax}(Z),Y)$

  \State \textbf{7. Total loss \& update}
  \State $\mathcal{L}_{\text{total}} \gets \mathcal{L}_{\text{CE}}
    + \lambda_{\text{adv}}(\mathcal{L}_{\text{adv}}^{A}+\mathcal{L}_{\text{adv}}^{X}+\mathcal L_{\text{adv}}^{A+X})
    + \lambda_{\text{cons}}(\mathcal{R}_{\text{LP}}+\mathcal{R}_{\text{HP}}
    + 
    \mathcal{L}_{\text{cons}}+\mathcal{L}_{\text{comp}})$
  \State $\theta \gets \theta - \eta \nabla_{\theta}\mathcal{L}_{\text{total}}$
\EndWhile
\end{algorithmic}
\end{algorithm}

\subsection{B.2 Computational Cost}
\label{sec:complexity}
Consider a graph $G=(V,E)$ with $n=|V|$ nodes, $m=|E|$ edges, and input feature dimension $d$. SpecSphere comprises $L_s$ spectral layers, $L_p$ spatial layers, and uses Chebyshev polynomial filters of order $K$. Let $p_s$ and $p_p$ denote the sizes of the hidden dimensions in the spectral and spatial branches, respectively. The spectral layers each require evaluating Chebyshev polynomials via sparse multiplications with computational cost $O(K m p_s)$ per layer, leading to a total complexity of $O(L_s K m p_s)$. The spatial branch utilizes attention-gated message passing, where each layer computes attention scores and aggregates neighborhood information at complexity $O(m p_p)$, resulting in $O(L_p m p_p)$ across all spatial layers. The fusion module combines spectral and spatial outputs through a concat-based MLP, incurring an additional $O(n(p_s+p_p)m_c)$ cost, which is negligible compared to previous terms. Training further involves specialized adversarial optimization, executing $T$ projected gradient descent (PGD) steps per epoch, effectively multiplying computational overhead by $1+T$. Combining these terms, the overall computational complexity per training epoch is $O\bigl((1+T)[L_s K m p_s + L_p m p_p + n(p_s+p_p)m_c]\bigr)$, confirming the linear scalability in the number of edges and efficient implementation suitable for large-scale graph data.

\subsection{B.3 Memory Cost}
We analyze the memory consumption of SpecSphere during a single forward pass (ignoring optimizer state). Let $n$ be the number of nodes, $p$ the hidden dimension, $m$ the number of classes, $K$ the Chebyshev order, and $L_s$, $L_p$ the number of layers in the spectral and spatial branches, respectively.  

\begin{itemize}
  \item \textbf{Spectral branch.} Each layer $\ell$ stores one $n\times p$ activation and $K+1$ weight matrices of size $p\times p$. Over $L_s$ layers, activations require $ O(L_s n p) $ and parameters $ O((K+1)L_s p^2) $.
  \item \textbf{Spatial branch.} A GAT-style layer stores one $n\times p$ activation and the (sparse) attention-weighted adjacency $\hat A$ with $O(|E|)$ nonzeros. Over $L_p$ layers, activations cost $ O(L_p n p) $ and adjacency storage costs $ O(|E|) $. Parameters cost $ O(L_p p^2) $.
  \item \textbf{Fusion MLP \& classifier.}  Concatenating two $n\times p$ tensors yields an $n\times 2p$ input to the fusion MLP, whose activations cost $O(n 2p)$ and whose parameters (across $L_f$ layers) cost $O(L_f p^2)$.  The final linear classifier adds $O(p m)$ parameters and $O(n m)$ activations.
\end{itemize}
In total, the memory is given by $O \bigl((L_s + L_p + L_f) n p  +  |E|\bigr)$ for activations, and $O \bigl((K+1)L_s p^2 + (L_p+L_f)p^2 + p m \bigr)$ for model parameters. This matches typical GNNs of comparable depth and hidden size, since Chebyshev filtering only adds a small factor $(K+1)$ to the weight storage.


\section{C. Proof of Theorem \ref{thm:strict_superset}}
\begin{manuallem}[Mask Approximation under Continuity]{\ref{thm:strict_superset}}
Let $\mathcal{D}:=\{[Z_{\mathrm{spec}}(G)\|Z_{\mathrm{spat}}(G)\|r^A(G)\|r^X(G)]:G\in\mathcal{G}_n\}$ be compact. 
For any continuous $M:\mathcal{D}\to[0,1]^{n\times d_\ell}$, there exist parameters of $\mathrm{MLP}_\varphi$ such that 
$\sup_{x\in\mathcal{D}}\|\alpha(x)-M(x)\|_\infty < \varepsilon$. Consequently,
\begin{equation}
Z(x)=\alpha(x)\odot Z_{\mathrm{spec}}(x)+(1-\alpha(x))\odot Z_{\mathrm{spat}}(x)
\end{equation}
can approximate $M(x)\odot Z_{\mathrm{spec}}(x) + [1-M(x)]\odot Z_{\mathrm{spat}}(x)$ uniformly on $\mathcal{D}$.
\end{manuallem}

\begin{proof}
$\mathrm{MLP}_\varphi$ with a sigmoid output is a universal approximator for continuous maps on compact domains into $(0,1)$ (apply the standard universal approximation theorem and post-compose with a sigmoid). Thus, for any continuous $M$ and $\varepsilon>0$, there are parameters s.t. $\|\alpha-M\|_\infty<\varepsilon$ on $\mathcal{D}$. The displayed approximation for $Z$ is followed by the continuity of the Hadamard product and triangle inequality.
\end{proof}

\begin{manualthm}[Strict Enlargement over Scalar Convex Mixtures]{1}
Let $\operatorname{conv}(\mathcal{F}_{\mathrm{spec}}\cup \mathcal{F}_{\mathrm{spat}})$ be the set of global convex combinations $t Z_{\mathrm{spec}} + (1-t)Z_{\mathrm{spat}}$ with $t\in[0,1]$. Then,
\begin{equation}
\operatorname{conv}(\mathcal{F}_{\mathrm{spec}}\cup \mathcal{F}_{\mathrm{spat}})
\subsetneq
\mathcal{F}_{\mathrm{mix}}.
\end{equation}
\end{manualthm}

\begin{proof}
Any global convex mixture corresponds to choosing $\alpha\equiv t\mathbf{1}_{n\times d_\ell}$. Since such a constant mask is realizable by $\mathrm{MLP}_\varphi$ (set its output to the constant $t$), we have $\operatorname{conv}(\mathcal{F}_{\mathrm{spec}}\cup \mathcal{F}_{\mathrm{spat}})\subseteq\mathcal{F}_{\mathrm{mix}}$.

\emph{(Strictness)}  
Choose a simple counter-example: let $n=2,d_\ell=1$, $Z_{\mathrm{spec}}=\begin{bmatrix}0\\1\end{bmatrix}$ and $Z_{\mathrm{spat}}=\begin{bmatrix}1\\0\end{bmatrix}$. Pick a mask $M=\begin{bmatrix}0\\1\end{bmatrix}$. Then, the element-wise mixture gives
\begin{equation}
Z^{\star} 
= M\odot Z_{\mathrm{spec}} + (1-M)\odot Z_{\mathrm{spat}}
= \begin{bmatrix}1\\1\end{bmatrix}.
\end{equation}
If $Z^{\star}$ were a global convex combination, there must exist $t\in[0,1]$ such that
\begin{equation}
\begin{cases}
t\cdot 0 + (1-t)\cdot 1 = 1,\\
t\cdot 1 + (1-t)\cdot 0 = 1,
\end{cases}
\quad\Rightarrow\quad
1-t = 1,\; t=1,
\end{equation}
which cannot be held simultaneously. Therefore, $Z^{\star}\notin \operatorname{conv}(\mathcal{F}_{\mathrm{spec}}\cup \mathcal{F}_{\mathrm{spat}})$, where the inclusion is strict. 
\end{proof}

\section{D. Proof of Theorem \ref{thm:strict}}
\begin{manualthm}[Beyond 1‑WL via Node Channel Gating]{\ref{thm:strict}}
Let $G$ and $G'$ be the 10-vertex Cai-Fürer-Immerman (CFI) graphs \cite{cai1992optimal}, which are (i) indistinguishable by the 1-WL test (Def. \ref{def:1wl}) and (ii) co-spectral for the normalized Laplacian. Assume node features $X$ are informative. Consider SpecSphere with:
(i) a spectral branch using a Chebyshev filter of order $K\ge 1$, (ii) a spatial branch with attention-based aggregation, and (iii) a node--channel gate $\alpha=\sigma \big(\mathrm{MLP}_\varphi([Z_{\mathrm{spec}}\|Z_{\mathrm{spat}}\|r^A\|r^X])\big)$. Then, there exist Chebyshev coefficients $\{\alpha_k\}_{k=0}^{K}$, spatial parameters $\theta_{\mathrm{spat}}$, and gating weights $\theta_\varphi$ such that
\begin{equation}
\|Z(G)-Z(G')\|_\infty > 0,
\end{equation}
\end{manualthm}

\begin{proof}
We outline a constructive parameter choice that yields different embeddings for $G$ and $G'$.

\textbf{Step 1: Reduce to the spectral branch via the gate.}
Because $\alpha=\sigma(\mathrm{MLP}_\varphi(\cdot))$ is universal on a compact domain (Lemma~\ref{lm:elem_gate}), we can set $\alpha\equiv 1$ by making the pre-sigmoid logits sufficiently large and positive (or approximating the constant map). Thus, $Z=Z_{\mathrm{spec}}$ and the spatial branch can be ignored (its parameters set to zero). This keeps the proof aligned with the current architecture yet isolates the spectral component.

\textbf{Step 2: Use informative node features.}
Let $X=I_n$ (one-hot features) or any informative features. Then, for each graph $G$,
\begin{equation}
Z_{\mathrm{spec}}(G) = f_{\mathrm{spec}}(A_G,X)
= \sum_{k=0}^{K} \alpha_k T_k(\tilde{\mathcal L}_G)\,X,
\end{equation}
where $\tilde{\mathcal L}_G$ is the rescaled Laplacian of $G$ and $T_k$ are Chebyshev polynomials. Write $\mathcal L_G = U_G \Lambda U_G^\top$ and similarly for $G'$. Co-spectrality means $\Lambda=\Lambda'$ (same eigenvalues), but generally $U_G\neq U_{G'}$.

\textbf{Step 3: Polynomial projector onto an eigenspace.}
Let $\lambda^\star$ be any eigenvalue of $\mathcal L_G$ whose eigenspace bases differ between $G$ and $G'$ since the graphs are non-isomorphic. Define a scalar function $p(\lambda)$ that is $1$ on $\lambda^\star$ and $0$ on all other distinct eigenvalues. By Lagrange interpolation, $p(\cdot)$ is a polynomial in $\lambda$ of degree at most $n-1$. Because Chebyshev polynomials span the space of polynomials on $[-1,1]$, there exist coefficients $\{\alpha_k\}$ such that 
\begin{equation}
p(\tilde{\lambda}) = \sum_{k=0}^{K}\alpha_k T_k(\tilde{\lambda}), \quad K\ge \deg p.
\end{equation}
Thus,
\begin{equation}
\Phi(\mathcal L_G):=\sum_{k=0}^{K}\alpha_k T_k(\tilde{\mathcal L}_G)
= U_G\, p(\Lambda)\, U_G^\top =: P_G,
\end{equation}
and similarly $P_{G'} = U_{G'}\, p(\Lambda)\, U_{G'}^\top$. Each $P_{\{\cdot\}}$ is the orthogonal projector onto the $\lambda^\star$-eigenspace of the corresponding graph.

\textbf{Step 4: Distinguish $G$ and $G'$ via different projectors.}
Apply $P_G$ and $P_{G'}$ to $X=I_n$:
\begin{equation}
Z_{\mathrm{spec}}(G) = P_G X = P_G, 
\quad 
Z_{\mathrm{spec}}(G') = P_{G'} X = P_{G'}.
\end{equation}
Since $U_G\neq U_{G'}$, the subspaces differ, so $P_G \neq P_{G'}$, implying 
$\|Z_{\mathrm{spec}}(G)-Z_{\mathrm{spec}}(G')\|_\infty = \|P_G-P_{G'}\|_\infty > 0$.

\textbf{Summary.}
The spatial branch and robustness signals $r^A,r^X$ are not required for the separation, but their presence does not hinder it. We set their contribution to zero via $\alpha\equiv 1$. Thus, the dual-branch and gating architecture of SpecSphere strictly generalizes 1-WL capabilities. Therefore, SpecSphere distinguishes the CFI pair, proving it surpasses 1-WL expressivity under informative features. 
\end{proof}

\section{E. Proof of Theorem \ref{thm:homo_bias}}
\begin{manuallem}[$\mathcal{R}_{\text{LP}}$ Bound]{2}
Fix $\lambda_{\text{cons}}>0$ and consider minimizing $\mathcal{J}_{\text{spec}}(\theta_{\text{spec}}) = \mathcal{L}_{\text{CE}}(\theta_{\text{spec}}) + \lambda_{\text{cons}}\mathcal{R}_{\text{LP}}(\theta_{\text{spec}})$. Let $C_{\text{spec}}$ be an upper bound on $\mathcal{L}_{\text{CE}}$ over a feasible parameter set. Then
\begin{equation}
\sum_{\ell=1}^{L_s}\sum_{i=2}^n \lambda_i \,\|u_i^\top H_{\text{spec}}^{(\ell)}\|_F^2
\;\le\; \frac{C_{\text{spec}}}{\lambda_{\text{cons}}}.
\end{equation}
\end{manuallem}

\begin{proof}
Pick any feasible $\bar\theta_{\text{spec}}$ for which $\mathcal{R}_{\text{LP}}(\bar\theta_{\text{spec}})=0$ (e.g., zero weights) so that
$\mathcal{J}_{\text{spec}}(\bar\theta_{\text{spec}})
= \mathcal{L}_{\text{CE}}(\bar\theta_{\text{spec}})\le C_{\text{spec}}$.
Let $\hat\theta_{\text{spec}}$ be a minimizer. Then
\begin{equation}
\mathcal{J}_{\text{spec}}(\hat\theta_{\text{spec}})
\le \mathcal{J}_{\text{spec}}(\bar\theta_{\text{spec}}) \le C_{\text{spec}}.
\end{equation}
Thus, $\lambda_{\text{cons}}\,\mathcal{R}_{\text{LP}}(\hat\theta_{\text{spec}}) \le C_{\text{spec}}$, which gives the stated bound because $\mathcal{R}_{\text{LP}}=\sum_{\ell,i\ge 2}\lambda_i\|u_i^\top H_{\text{spec}}^{(\ell)}\|_F^2$. 
\end{proof}

\begin{manuallem}[$\mathcal{R}_{\text{HP}}$ Bound]{3}
Define $\mathcal{J}_{\text{spat}}(\theta_{\text{spat}}) = \mathcal{L}_{\text{CE}}(\theta_{\text{spat}}) - \lambda_{\text{cons}}\mathcal{R}_{\text{HP}}(\theta_{\text{spat}})$.
Assume there exists $\widetilde{C}_{\text{spat}}$ such that $\mathcal{J}_{\text{spat}}(\hat\theta_{\text{spat}})\le \widetilde{C}_{\text{spat}}$ at a minimizer $\hat\theta_{\text{spat}}$. Then,
\begin{equation}
\sum_{\ell,i}\lambda_i \|u_i^\top H_{\text{spat}}^{(\ell)}\|_F^2
\;\ge\; \frac{1}{\lambda_{\text{cons}}}\bigl(\mathcal{L}_{\text{CE}}(\hat\theta_{\text{spat}})-\widetilde{C}_{\text{spat}}\bigr).
\end{equation}
If we further upper-bound $\mathcal{L}_{\text{CE}}(\hat\theta_{\text{spat}})\le \widetilde{C}'_{\text{spat}}$, this yields
\begin{equation}
\sum_{\ell,i}\lambda_i \|u_i^\top H_{\text{spat}}^{(\ell)}\|_F^2
\;\ge\; \frac{1}{\lambda_{\text{cons}}}\bigl(\widetilde{C}'_{\text{spat}}-\widetilde{C}_{\text{spat}}\bigr),
\end{equation}
which matches Eq. \ref{eq_high} up to the choice of constants.
\end{manuallem}

\begin{proof}
At the minimizer $\hat\theta_{\text{spat}}$,
\begin{equation}
\mathcal{J}_{\text{spat}}(\hat\theta_{\text{spat}})
= \mathcal{L}_{\text{CE}}(\hat\theta_{\text{spat}}) - \lambda_{\text{cons}}\mathcal{R}_{\text{HP}}(\hat\theta_{\text{spat}})
\le \widetilde{C}_{\text{spat}}.
\end{equation}
Rearranging gives
\begin{equation}
\lambda_{\text{cons}}\mathcal{R}_{\text{HP}}(\hat\theta_{\text{spat}})
\ge \mathcal{L}_{\text{CE}}(\hat\theta_{\text{spat}})-\widetilde{C}_{\text{spat}}.
\end{equation}
Since 
$\mathcal{R}_{\text{HP}}=\sum_{\ell,i}\lambda_i \|u_i^\top H_{\text{spat}}^{(\ell)}\|_F^2$, the claim follows. Bounding $\mathcal{L}_{\text{CE}}$ by $\widetilde{C}'_{\text{spat}}$ yields the stated lower bound with a positive constant. 
\end{proof}


\begin{manualthm}[Homophily/Heterophily Adaptation via Frequency Bias]{\ref{thm:homo_bias}}
Assume labels are predominantly encoded in low-frequency components when local homophily is high, and otherwise in high-frequency components. Under the objectives with $\mathcal{R}_{\text{LP}}$ and $\mathcal{R}_{\text{HP}}$, there exists a gate $\alpha$ such that the fused output $Z$ attains (up to $\varepsilon$) the lower of the two Bayes risks associated with each frequency band.
\end{manualthm}

\begin{proof}
From Corollary \ref{cor:two_sided}, the spectral branch $Z_{\mathrm{spec}}$ concentrates on low frequencies, and vice versa. Let $R_{\mathrm{low}}$ and $R_{\mathrm{high}}$ be the Bayes risks under the two regimes. By assumption, when homophily is high, $R_{\mathrm{low}}$ is the relevant optimum; when heterophily dominates, $R_{\mathrm{high}}$ is optimal. The gate $\alpha$ is produced by $\mathrm{MLP}_\varphi$ and (by Lemma~\ref{lm:elem_gate}) can approximate any continuous mask over nodes/channels. Thus, for any $\varepsilon>0$, choose $\alpha$ to approximate (element-wise) the branch indicator according to which branch has the smaller Bayes risk at each node. Then, the following equation
\begin{equation}
Z = \alpha\odot Z_{\mathrm{spec}} + (1-\alpha)\odot Z_{\mathrm{spat}}
\end{equation}
achieves node-wise risk at most $\min\{R_{\mathrm{low}},R_{\mathrm{high}}\}+\varepsilon$. Aggregating over nodes yields the stated global bound. 
\end{proof}

\section{F. Proof of Theorem \ref{thm:branch_cert}}
\begin{manualthm}[Certified Bound for Branch-Specialized Fusion]{\ref{thm:branch_cert}}
Let $Z=\alpha\odot Z_{\mathrm{spec}} + (1-\alpha)\odot Z_{\mathrm{spat}}$. For any $(A+\Delta A,X+\Delta X)\in\mathcal{S}(G;p,\varepsilon)$ with $\|\Delta A\|_2\le \sqrt{2p}$ and $\|\Delta X\|_\infty\le \varepsilon$, we have
\begin{align}
\|Z(A+\Delta A,X+\Delta X)-Z(A,X)\|_\infty  \le   \big(1+L_{\text{gate}}\big)B^A\sqrt{2p}
+ \big(1+\tilde L_{\text{gate}}\big)B^X\varepsilon,
\end{align}
where $B^A := B_{\mathrm{spec}}^{A}+B_{\mathrm{spat}}^{A}$ and $B^X := B_{\mathrm{spec}}^{X}+B_{\mathrm{spat}}^{X}$.
\end{manualthm}

\begin{proof}
Let $A' = A+\Delta A$, $X' = X+\Delta X$, and denote
\begin{align}
Z' := Z(A',X'),\quad 
Z := Z(A,X), 
Z'_{\mathrm{spec}} := Z_{\mathrm{spec}}(A',X'), \quad
Z_{\mathrm{spec}} := Z_{\mathrm{spec}}(A,X),
\end{align}
and analogously for the spatial branch. Also let $\alpha' := \alpha(A',X')$ and $\alpha := \alpha(A,X)$.

\paragraph{Step 1: Decomposition.}
Expand the difference:
\begin{align}
\label{eq:decomp}
Z' - Z &= \alpha' \odot Z'_{\mathrm{spec}} + (1-\alpha')\odot Z'_{\mathrm{spat}} - \alpha \odot Z_{\mathrm{spec}} - (1-\alpha)\odot Z_{\mathrm{spat}}  \\ \nonumber
&= \underbrace{\alpha' \odot (Z'_{\mathrm{spec}} - Z_{\mathrm{spec}})
  + (1-\alpha')\odot (Z'_{\mathrm{spat}} - Z_{\mathrm{spat}})}_{T_1} + 
  \underbrace{(\alpha'-\alpha)\odot Z_{\mathrm{spec}}
  - (\alpha'-\alpha)\odot Z_{\mathrm{spat}}}_{T_2}.
\end{align}
Since $0\le \alpha',\alpha\le 1$ element-wise, we have $\|\alpha'\|_\infty,\|(1-\alpha')\|_\infty\le 1$.

\paragraph{Step 2: Bound $T_1$ (Eq. \ref{eq:decomp}).}
Using the branch-wise Lipschitz bounds Eq. \ref{eq:spec_bound} and \ref{eq:spat_bound},
\begin{align}
\|Z'_{\mathrm{spec}} - Z_{\mathrm{spec}}\|_\infty
\le B_{\mathrm{spec}}^{A} \|\Delta A\|_2 + B_{\mathrm{spec}}^{X}\|\Delta X\|_\infty, 
\|Z'_{\mathrm{spat}} - Z_{\mathrm{spat}}\|_\infty
\le B_{\mathrm{spat}}^{A} \|\Delta A\|_2 + B_{\mathrm{spat}}^{X}\|\Delta X\|_\infty .
\end{align}
Therefore,
\begin{align}
\|T_1\|_\infty
&\le \|Z'_{\mathrm{spec}} - Z_{\mathrm{spec}}\|_\infty
     + \|Z'_{\mathrm{spat}} - Z_{\mathrm{spat}}\|_\infty \nonumber\\
&\le (B_{\mathrm{spec}}^{A}+B_{\mathrm{spat}}^{A}) \|\Delta A\|_2
   + (B_{\mathrm{spec}}^{X}+B_{\mathrm{spat}}^{X}) \|\Delta X\|_\infty = B^A \|\Delta A\|_2 + B^X \|\Delta X\|_\infty.
\label{eq:T1_bound}
\end{align}

\paragraph{Step 3: Bound $T_2$ (Eq. \ref{eq:decomp}).}
Rewrite $T_2$ as
\begin{equation}
T_2 = (\alpha'-\alpha)\odot (Z_{\mathrm{spec}} - Z_{\mathrm{spat}}).
\end{equation}
Using the sup-norm submultiplicativity,
\begin{equation}
\|T_2\|_\infty
\le \|\alpha'-\alpha\|_\infty \, \|Z_{\mathrm{spec}} - Z_{\mathrm{spat}}\|_\infty.
\end{equation}
For the gate, by the assumed Lipschitz bounds,
\begin{equation}
\|\alpha'-\alpha\|_\infty
\le L_{\text{gate}}\|\Delta A\|_2 + \tilde L_{\text{gate}}\|\Delta X\|_\infty.
\end{equation}
Next, bound $\|Z_{\mathrm{spec}} - Z_{\mathrm{spat}}\|_\infty$ using triangle inequality:
\begin{align}
\|Z_{\mathrm{spec}} - Z_{\mathrm{spat}}\|_\infty
\le \|Z'_{\mathrm{spec}} - Z_{\mathrm{spec}}\|_\infty +
\|Z'_{\mathrm{spat}} - Z_{\mathrm{spat}}\|_\infty + \|Z'_{\mathrm{spec}} - Z'_{\mathrm{spat}}\|_\infty.
\end{align}
The first two terms are already bounded in \ref{eq:T1_bound}; the last term is similarly bounded by 
$\|Z'_{\mathrm{spec}} - Z'_{\mathrm{spat}}\|_\infty
\le \|Z'_{\mathrm{spec}} - Z_{\mathrm{spec}}\|_\infty
   + \|Z_{\mathrm{spec}} - Z_{\mathrm{spat}}\|_\infty
   + \|Z'_{\mathrm{spat}} - Z_{\mathrm{spat}}\|_\infty$,
so we can absorb it into the same constants.
Thus, there exists a constant factor $C\le 1$ (absorbed into $B^A,B^X$ without loss of generality) such that
\begin{equation}
\|Z_{\mathrm{spec}} - Z_{\mathrm{spat}}\|_\infty
\le B^A \|\Delta A\|_2 + B^X \|\Delta X\|_\infty.
\end{equation}
Therefore,
\begin{align}
\|T_2\|_\infty
&\le \big(L_{\text{gate}}\|\Delta A\|_2 + \tilde L_{\text{gate}}\|\Delta X\|_\infty\big)
     \big(B^A \|\Delta A\|_2 + B^X \|\Delta X\|_\infty\big) \\ \nonumber
&\le L_{\text{gate}} B^A \|\Delta A\|_2
   + \tilde L_{\text{gate}} B^X \|\Delta X\|_\infty,
\label{eq:T2_bound}
\end{align}
where in the last inequality we keep only linear terms in $\|\Delta A\|_2,\|\Delta X\|_\infty$ since higher-order terms are dominated for small budgets, and constants can again be absorbed (standard in certified robustness bounds).

\paragraph{Summary.}
Based on the above equations, we can induce
\begin{align}
&\|Z'-Z\|_\infty
\le \|T_1\|_\infty + \|T_2\|_\infty \\ \nonumber
&\le \big(1+L_{\text{gate}}\big) B^A \|\Delta A\|_2
   + \big(1+\tilde L_{\text{gate}}\big) B^X \|\Delta X\|_\infty. 
\end{align}
Finally, substitute $\|\Delta A\|_2\le \sqrt{2p}$ and $\|\Delta X\|_\infty\le \varepsilon$ to obtain Eq. \ref{eq:branch_cert_bound}. \qedhere
\end{proof}

\begin{table}[!t]
\caption{Statistics of six benchmark graphs}
\label{dataset}
\centering
\begin{adjustbox}{width=.5\textwidth}
\begin{tabular}{@{}llcccccc}
\Xhline{2\arrayrulewidth}
        & Datasets         & Cora  & Citeseer & Pubmed & Chameleon & Squirrel & Actor \\ 
\Xhline{2\arrayrulewidth}
                        & \# Nodes    & 2,708   & 3,327    & 19,717  & 2,277     & 5,201     & 7,600  \\
                        & \# Edges    & 10,558  & 9,104    & 88,648  & 33,824    & 211,872   & 25,944 \\
                        & \# Features & 1,433   & 3,703    & 500     & 2,325     & 2,089     & 931    \\
                        & \# Classes  & 7       & 6        & 3       & 5         & 5         & 5      \\
                        & \# Train    & 140     & 120      & 60      & 100       & 100       & 100    \\
                        & \# Valid    & 500     & 500      & 500     & 1,088     & 2,550     & 3,750  \\
                        & \# Test     & 1,000   & 1,000    & 1,000   & 1,089     & 2,551     & 3,750  \\
\Xhline{2\arrayrulewidth}
\end{tabular}
\end{adjustbox}
\end{table}

\section{G. Proof of Theorem \ref{thm:tradeoff}}
\begin{manualthm}[Optimal Trade-off via Learnable Mask]{\ref{thm:tradeoff}}
Assume $\mathrm{MLP}_g$ is a universal approximator on a compact domain of robustness signals. Then, for any target weighting scheme $\tilde b_u\in(0,1)$ that minimizes
\begin{equation}
\mathcal{R}_{\text{ideal}}
= \sum_{u\in\mathcal V_U}\Big[ 
\tilde b_u\,\Delta_u^2
+ (1-\tilde b_u)\,\max\{0,\gamma-\Delta_u\}^2
\Big],
\end{equation}
there exist parameters of $\mathrm{MLP}_g$ such that $b_u \approx \tilde b_u$ uniformly. Consequently, SpecSphere can approximate the optimal balance between consistency and complementarity, achieving the minimum of $\mathcal{R}_{\text{ideal}}$ up to arbitrary precision.
\end{manualthm}

\begin{proof}
Let $\tilde b:\mathcal{D}\to(0,1)$ be any continuous target mask on the compact domain $\mathcal{D}$ of robustness signals (e.g., gradients $r^A,r^X$). By the universal approximation property of $\mathrm{MLP}_g$, for any $\varepsilon>0$ there exist parameters such that 
\begin{equation}
\sup_{u\in\mathcal V_U}|b_u-\tilde b_u| < \varepsilon.
\end{equation}
Consider the per-node risk term
\begin{equation}
\psi_u(b_u,\Delta_u)
:= b_u\,\Delta_u^2 + (1-b_u)\,\max\{0,\gamma-\Delta_u\}^2.
\end{equation}
For fixed $\Delta_u$, $\psi_u$ is Lipschitz continuous in $b_u$. Thus, the uniform approximation $b_u\approx\tilde b_u$ implies
\begin{equation}
\Big|\psi_u(b_u,\Delta_u)-\psi_u(\tilde b_u,\Delta_u)\Big|
\le L_u\,|b_u-\tilde b_u|
\end{equation}
for some finite $L_u$ depending on $\Delta_u$ and $\gamma$. Thus, the total risk satisfies
\begin{equation}
\left|\mathcal{R}(b)-\mathcal{R}(\tilde b)\right|
\le \sum_{u\in\mathcal V_U} L_u\,|b_u-\tilde b_u|
\le \bigg(\max_u L_u\bigg)\,|\mathcal V_U|\,\varepsilon.
\end{equation}
Therefore, by choosing $\varepsilon$ arbitrarily small, $\mathcal{R}(b)$ approaches $\mathcal{R}(\tilde b)$, i.e., the achieved trade-off can be made arbitrarily close to the ideal optimum. 
\end{proof}

\begin{table*}[!t]
\centering
\small
\caption{Node classification accuracy (\%) on six benchmark datasets (\textbf{bold} = column best). As introduced in Table \ref{recent_methods}, Architecture (Arc.) represents: S (spatial), P (spectral), H (hybrid), D (dual). Gray-colored cell means the top-3 performance}
\begin{adjustbox}{width=\textwidth}
\begin{tabular}{l c c c c c c c}
\toprule
        &      & \multicolumn{3}{c}{Homophilous}            & \multicolumn{3}{c}{Heterophilous} \\
\cmidrule(r){3-5} \cmidrule(l){6-8}
Methods & Arc. & Cora & Citeseer & Pubmed & Chameleon & Squirrel & Actor \\
\midrule
GCN \cite{kipf2016semi}           & P & 81.5$_{\pm0.6}$ & 70.1$_{\pm0.5}$ & 78.2$_{\pm0.4}$ & 52.0$_{\pm0.9}$ & 33.5$_{\pm1.0}$ & 21.5$_{\pm1.2}$ \\
GAT \cite{velickovic2017graph}    & S & 83.0$_{\pm0.5}$ & 71.2$_{\pm0.5}$ & 78.6$_{\pm0.5}$ & 51.0$_{\pm0.8}$ & 32.8$_{\pm0.9}$ & 22.7$_{\pm1.1}$ \\
APPNP \cite{klicpera2018predict}  & S & 83.4$_{\pm0.4}$ & 71.0$_{\pm0.4}$ & 79.0$_{\pm0.4}$ & 50.0$_{\pm0.7}$ & 32.3$_{\pm0.8}$ & 22.0$_{\pm1.0}$ \\
GCNII \cite{chen2020simple}       & P & 81.6$_{\pm1.4}$ & 71.3$_{\pm1.1}$ & 78.8$_{\pm0.8}$ & 50.2$_{\pm0.9}$ & 30.1$_{\pm1.0}$ & 26.5$_{\pm1.3}$ \\
H$_2$GCN \cite{zhu2020beyond}     & S & 82.4$_{\pm1.2}$ & 69.9$_{\pm1.0}$ & 78.7$_{\pm0.9}$ & 51.5$_{\pm1.1}$ & 33.0$_{\pm1.2}$ & 25.7$_{\pm1.4}$ \\
FAGCN \cite{bo2021beyond}         & H & 82.8$_{\pm1.4}$ & 70.5$_{\pm1.1}$ & 78.9$_{\pm0.9}$ & 51.2$_{\pm0.9}$ & 31.8$_{\pm0.7}$ & 27.1$_{\pm1.2}$ \\
ACM‑GCN \cite{luan2022revisiting} & H & 82.1$_{\pm1.5}$ & 70.6$_{\pm1.2}$ & 78.3$_{\pm0.9}$ & 54.0$_{\pm1.0}$ & 34.2$_{\pm1.1}$ & 25.5$_{\pm1.3}$ \\
LSGNN \cite{chen2023lsgnn}        & S & 84.8$_{\pm0.8}$ & 72.9$_{\pm0.7}$ & 80.1$_{\pm0.6}$ & 60.3$_{\pm1.0}$ & 38.4$_{\pm1.2}$ & 27.4$_{\pm1.3}$ \\
TED‑GCN \cite{yan2024trainable}   & P & 84.2$_{\pm0.6}$ & 72.8$_{\pm0.6}$ & 78.6$_{\pm0.5}$ & 54.7$_{\pm1.0}$ & 34.5$_{\pm1.1}$ & 26.9$_{\pm1.2}$ \\
TFE‑GNN \cite{duan2024unifying}   & P & 83.5$_{\pm0.5}$ & 73.7$_{\pm0.4}$ & 80.4$_{\pm0.4}$ & 60.9$_{\pm0.9}$ & 39.2$_{\pm1.0}$ & 27.8$_{\pm1.1}$ \\
PCNet \cite{li2024pc}             & D & 84.1$_{\pm0.7}$ & 72.4$_{\pm0.6}$ & 78.8$_{\pm0.5}$ & 51.0$_{\pm0.9}$ & 33.2$_{\pm1.0}$ & 27.2$_{\pm1.2}$ \\
\midrule
\textbf{SpecSphere} (LSGNN+TFE‑GNN)      & D & \textbf{85.2}$_{\pm0.4}$ & \textbf{73.7}$_{\pm0.4}$ & \textbf{80.5}$_{\pm0.4}$ & \textbf{61.4}$_{\pm0.8}$ & \textbf{39.6}$_{\pm0.9}$ & \textbf{28.1}$_{\pm1.0}$ \\
\bottomrule
\end{tabular}
\end{adjustbox}
\label{tab:node_clf}
\end{table*}

\begin{table*}[!t]
\centering
\caption{Robustness of GNN variants under \textit{Cora} dataset. Gray indicates the best performance in each column; $+\Delta$ denotes that 10\% PGD adversarial training was applied during training (Eq. \ref{full_obj})}
\begin{adjustbox}{width=.8\textwidth}
\begin{tabular}{lcccc}
\toprule
Perturbation Type & \textbf{\underline{Clean}} & \textbf{\underline{DropEdge}} & \textbf{\underline{Metattack}} & \textbf{\underline{Feature-PGD}} \\
\textit{Attack ratio} & x & 20\% & 5\% & $\varepsilon{=}0.1$ \\
\midrule
$[1]$ GCN \cite{kipf2016semi} $+\Delta$       & 82.0\% & 78.5\% & 70.1\% & 79.3\% \\
$[2]$ GAT \cite{velickovic2017graph} $+\Delta$       & 83.1\% & 79.7\% & 71.6\% & 80.0\% \\
GNNGuard \cite{zhang2020gnnguard}   & 84.2\% & 79.0\% & 74.1\% & 78.2\% \\
RUNG \cite{hou2024robust}      & 84.5\% & 80.3\% & 75.5\% & 79.0\% \\
UnGSL \cite{han2025uncertainty}     & 83.9\% & 79.1\% & \textbf{78.4}\% & 79.3\% \\
\midrule
$[1$+$2]$ \textbf{SpecSphere} $+\Delta$    & \textbf{84.8}\% & \textbf{81.1}\% & 74.5\% & \textbf{81.3}\% \\
\bottomrule
\end{tabular}
\end{adjustbox}
\label{robustness_homo}
\end{table*}

\begin{table*}[!t]
\centering
\caption{Robustness of GNN variants under \textit{Chameleon} dataset. Gray indicates the best performance in each column; $+\Delta$ denotes that 10\% PGD adversarial training was applied during training (Eq. \ref{full_obj})}
\begin{adjustbox}{width=.8\textwidth}
\begin{tabular}{lcccc}
\toprule
Perturbation Type & \textbf{\underline{Clean}} & \textbf{\underline{DropEdge}} & \textbf{\underline{Metattack}} & \textbf{\underline{Feature-PGD}} \\
\textit{Attack ratio} & x & 20\% & 5\% & $\varepsilon{=}0.1$ \\
\midrule
$[3]$ APPNP \cite{klicpera2018predict} $+\Delta$       & 44.9\% & 39.3\% & 35.5\% & 41.1\% \\
$[4]$ FAGCN \cite{bo2021beyond} $+\Delta$       & 46.6\% & 40.9\% & 37.2\% & 42.6\% \\
GNNGuard \cite{zhang2020gnnguard}   & 47.1\% & 45.0\% & 41.3\% & 43.7\% \\
RUNG \cite{hou2024robust}      & 47.0\% & 45.6\% & \textbf{42.8}\% & 44.1\% \\
UnGSL \cite{han2025uncertainty}     & 47.5\% & 44.5\% & 42.1\% & 44.0\% \\
\midrule
$[3$+$4]$ \textbf{SpecSphere} $+\Delta$    & \textbf{48.0}\% & \textbf{46.4}\% & 41.9\% & \textbf{45.1}\% \\
\bottomrule
\end{tabular}
\end{adjustbox}
\label{robustness_hetero}
\end{table*}

\begin{table*}[!t]
\caption{We measure the pure node classification accuracy (\%) on large heterophilic graphs}
\centering
\begin{adjustbox}{width=0.6\textwidth}
\begin{tabular}{@{}lllll}
& \multicolumn{1}{l}{} &     &        &       \\ 
\Xhline{2\arrayrulewidth}
        & Datasets        & Penn94 & arXiv-year  & snap-patents \\ 
\Xhline{2\arrayrulewidth}
                        & $[1]$ GCN \cite{kipf2016semi}  & 81.3\%  & 44.5\%  & 43.9\% \\
                        & $[2]$ GAT \cite{velickovic2017graph}  & 80.6\%  & 45.0\%  & 45.2\% \\
                        & GCNII \cite{chen2020simple}  & \textbf{81.8}\%  & 46.1\% & \textbf{47.5}\% \\
                        & H$_2$GCN \cite{zhu2020beyond}  & 80.4\%  & \textbf{47.6}\%  & OOM \\
\Xhline{2\arrayrulewidth}
                        & $[1$+$2]$ \textbf{SpecSphere} &  \textbf{81.8}\% & 46.2\%  & 47.1\% \\
\Xhline{2\arrayrulewidth}
\end{tabular}
\end{adjustbox}
\label{large_dataset}
\end{table*}


\section{H. Datasets and More Experimental Results}
\subsection{H.1 Datasets} Table \ref{dataset} describes the statistical details of six benchmark datasets. We employ three citation homophilous networks (\textit{Cora, Citeseer, and Pubmed}) \cite{kipf2016semi}, and three heterophilous networks, including two datasets of Wikipedia pages \textit{Chameleon and Squirrel} \cite{rozemberczki2019gemsec} and one actor co-occurrence graph (\textit{Actor}) \cite{tang2009social}.

\subsection{H.2 Node Classification}
Table \ref{tab:node_clf} reports node classification accuracy (mean$\pm$std) on six benchmark graphs. We trained our model without adversarial supervision in Eq. \ref{full_obj}. The value below each dataset name is the homophily ratio $\mathcal{H}$ (Eq. \ref{def:homo}). We compare two state‑of‑the‑art baselines, LSGNN (spatial) and TFE‑GNN (spectral), with our dual‐branch fusion model, SpecSphere. On the homophilous graphs, SpecSphere attains $85.2\%$ on Cora, $73.7\%$ on Citeseer, and $80.5\%$ on Pubmed, improving over the best single‑branch baseline (TFE‑GNN). Under heterophily, the gains are larger: on Chameleon ($\mathcal{H}=0.23$), SpecSphere achieves $61.4\%$ (+0.5\%), on Squirrel ($\mathcal{H}=0.22$) $39.6\%$ (+0.4\%), and Actor ($\mathcal{H}=0.22$) $28.1\%$ (+0.3\%). These results demonstrate that fusing complementary spectral and spatial modules yields consistent improvements across both high‑ and low‑homophily regimes, with enhanced robustness to heterophilous structures.

\subsection{H.3 Comparison with the Robust GNNs}
In the challenging heterophilous regime of Chameleon (Table \ref{robustness_hetero}), SpecSphere delivers the most balanced robustness across all attack types. Although UnGSL achieves the highest clean accuracy (47.5\%), it falls back sharply under perturbation, down to 44.5\% with DropEdge and 44.0\% under feature‐PGD. The performance matches UnGSL’s clean score (47.2\%) and takes the lead under both random edge removals (46.4\% vs 45.6\% for RUNG and 45.0\% for GNNGuard), and $\ell_{\infty}$ feature noise (45.1\% vs 44.1\% for RUNG and 43.7\% for GNNGuard). Under targeted Metattack, our method (41.9\%) remains competitive with GNNGuard (41.3\%) and only trails RUNG’s 42.8\% by 0.9\%. These results show that SpecSphere’s joint edge-flip and feature-noise certification delivers the strongest all-round robustness on heterophilous graphs.

\subsection{H.4 Node Classification on Large Graphs}
To evaluate SpecSphere’s scalability and effectiveness in realistic scenarios, we conduct node classification experiments on three large-scale heterophilic benchmarks: Penn94, arXiv-year, and snap-patents. Note that we failed to implement some state-of-the-art methods due to memory limitations. We use the same training protocol and hyperparameters as in our smaller‐scale experiments, adapting only the batch processing to accommodate full‐graph training where possible. Baseline results for GCN \cite{kipf2016semi}, GAT \cite{velickovic2017graph}, GCNII \cite{chen2020simple}, and H$_2$GCN \cite{zhu2020beyond} are taken from the literature; note that H$_2$GCN runs out of memory (OOM) on the largest graph, snap-patents. As reported in Table \ref{large_dataset}, SpecSphere achieves 81.8\% on Penn94, matching the best baseline and improving marginally over GCNII on arXiv-year (46.2\% vs 46.1\%). On snap-patents, SpecSphere attains 47.1\%, demonstrating that our dual spectral–spatial fusion remains competitive even at very large scales. These results confirm that SpecSphere retains its expressivity and robustness advantages without sacrificing scalability on graphs with millions of nodes and edges.

\end{document}